\documentclass{article} 

\usepackage{amsmath,amsfonts,bm}

\def\eqref#1{equation~\ref{#1}}

\def\1{\bm{1}}

\DeclareMathAlphabet{\mathsfit}{\encodingdefault}{\sfdefault}{m}{sl}
\SetMathAlphabet{\mathsfit}{bold}{\encodingdefault}{\sfdefault}{bx}{n}

\usepackage[dvipsnames]{xcolor}
\usepackage{hyperref}
\hypersetup{
    colorlinks=true,
    citecolor=MidnightBlue,
    linkcolor=OrangeRed,
    urlcolor=MidnightBlue
}
\usepackage[capitalise,nameinlink]{cleveref} 
\usepackage{iclr2025_conference,times}
\usepackage{url}
\usepackage{graphicx}
\usepackage{color}
\usepackage{colortbl}
\usepackage{subcaption}
\usepackage{enumitem}
\usepackage{booktabs}
\usepackage{makecell}
\usepackage{multicol}
\usepackage{multirow}
\usepackage{wrapfig}
\usepackage{algorithm}
\usepackage{algpseudocode}

\definecolor{unidrive_blue}{RGB}{136, 140, 188}

\title{UniDrive: Towards Universal Driving \\ Perception Across Camera Configurations}

\author{Ye Li$^1$\thanks{Work done while visiting UC Berkeley.} \quad 
Wenzhao Zheng$^2$\thanks{Corresponding author.} \quad 
Xiaonan Huang$^1$ \quad 
Kurt Keutzer$^2$ \\
$^1$University of Michigan, Ann Arbor \quad 
$^2$University of California, Berkeley\\
\url{https://wzzheng.net/UniDrive}\\
}

\iclrfinalcopy 

\begin{document}

\maketitle


\begin{abstract}

Vision-centric autonomous driving has demonstrated excellent performance with economical sensors. As the fundamental step, 3D perception aims to infer 3D information from 2D images based on 3D-2D projection. This makes driving perception models susceptible to sensor configuration (e.g., camera intrinsics and extrinsics) variations. However, generalizing across camera configurations is important for deploying autonomous driving models on different car models. In this paper, we present \textbf{UniDrive}, a novel framework for vision-centric autonomous driving to achieve universal perception across camera configurations. We deploy a set of unified virtual cameras and propose a ground-aware projection method to effectively transform the original images into these unified virtual views. We further propose a virtual configuration optimization method by minimizing the expected projection error between original and virtual cameras. The proposed virtual camera projection can be applied to existing 3D perception methods as a plug-and-play module to mitigate the challenges posed by camera parameter variability, resulting in more adaptable and reliable driving perception models. To evaluate the effectiveness of our framework, we collect a dataset on CARLA by driving the same routes while only modifying the camera configurations. Experimental results demonstrate that our method trained on one specific camera configuration can generalize to varying configurations with minor performance degradation. 

\end{abstract}

\section{Introduction}

Vision-centric autonomous driving has gained significant traction~\citep{wang2023exploring} due to its ability to deliver high-performance perception using economical sensors like cameras. At the core of this approach lies 3D perception~\citep{liu2022petr}, which reconstructs 3D spatial information from 2D images via 2D-3D lift transform~\citep{philion2020lift}. This transform is critical for enabling vehicles to understand their environment, detect objects, and navigate safely. Previous works~\citep{huang2021bevdet,xie2204m2bev,reading2021categorical,li2022bevformer,zhou2022cross,zeng2024hardness,lu2022learning,huang2022bevdet4d,liu2023petrv2} have achieved remarkable 3D perception ability by utilizing Bird's Eye View (BEV) representations to process 2D-3D lift. Recently, many vision-based 3D occupancy prediction methods~\citep{huang2023tri,wei2023surroundocc,huang2024selfocc,huang2024gaussianformer,zhao2024lowrankocc} further improved the understanding of dynamic and cluttered driving scenes, pushing the boundaries of the research domain. As a result, vision-based systems have become one of the primary solutions for scalable autonomous driving.

\begin{figure}[t]
    \centering
    \begin{subfigure}[b]{0.48\textwidth}
        \centering
        \includegraphics[width=\textwidth]{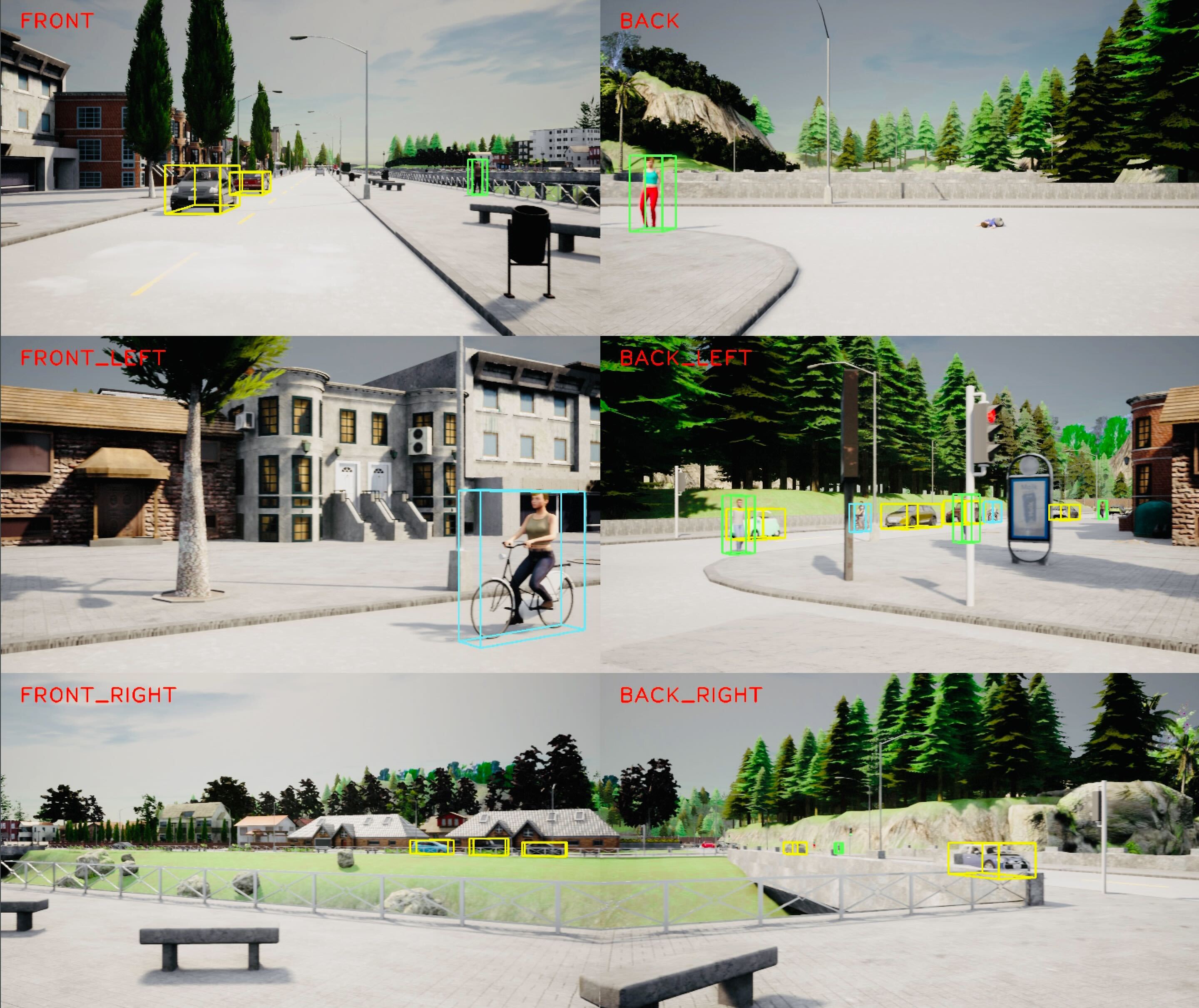}
        \caption{Deploy on \emph{same} camera configurations: \textcolor{blue}{Succeed!}}
        \label{fig:succeed}
    \end{subfigure}
    \hfill
    \begin{subfigure}[b]{0.48\textwidth}
        \centering
        \includegraphics[width=\textwidth]{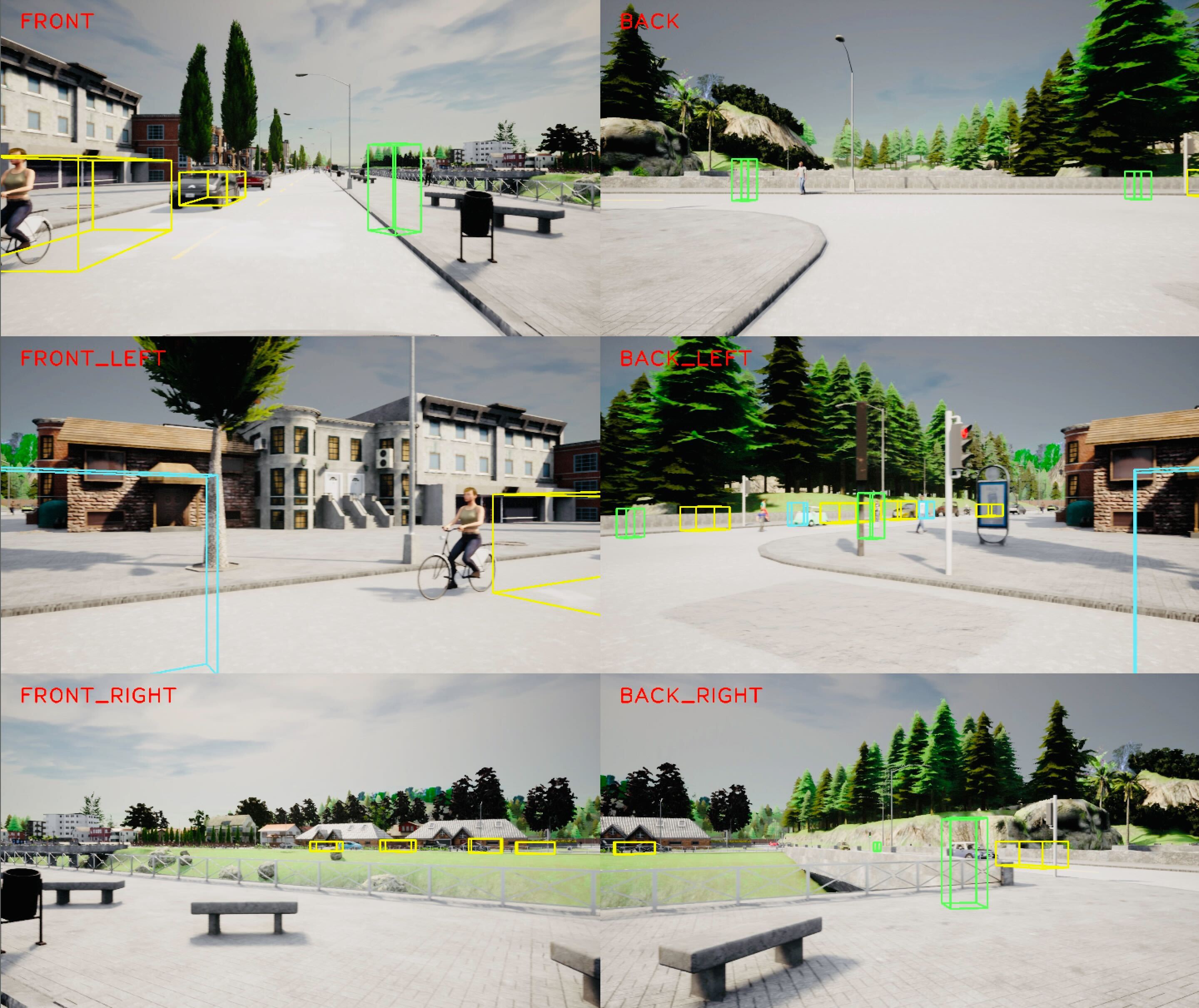}
        \caption{Deploy on \emph{different} camera configurations: \textcolor{red}{Fail!}}
        \label{fig:fail}
    \end{subfigure}
    \caption{Comparison of deploying perception models on the same and distinct configurations.}
    \label{fig:com}
    \vspace{-2mm}
\end{figure}

Despite the exciting development of the state-of-the-art vision-based autonomous driving~\citep{liu2024ray,zong2023temporal,zheng2024genad,zheng2023occworld,hu2023uniAD,jiang2023vad}, a critical limitation still remains: the sensitivity of these models to variations in camera configurations, including intrinsics and extrinsics. Autonomous driving models typically rely on well-calibrated sensor setups, and even slight deviations in camera parameters across different vehicles or platforms can significantly degrade performance~\citep{wang2023towards}. As illustrated in Figure~\ref{fig:succeed}, this lack of robustness to sensor variability makes it challenging to transfer perception models between different vehicle platforms without extensive retraining or manual adjustment.
This variation necessitates training separate models for each vehicle, which consumes a significant amount of computational resources. Thus, achieving generalization across camera configurations is essential for the practical deployment of vision-centric autonomous driving.

In this paper, we address two key questions surrounding generalizable driving perception: \textit{1) How can we construct a unified framework that enables perception models to generalize across different multi-camera parameters? 2) How can we further optimize the generalization of perception models to ensure robust performance across varying multi-camera configurations?}

To achieve this, we introduce \textbf{UniDrive}, a novel framework designed to address the challenge of generalizing perception models across multi-camera configurations. This framework deploys a set of unified virtual camera spaces and leverages a ground-aware projection method to transform original camera images into these unified virtual views. Additionally, we propose a virtual configuration optimization strategy that minimizes the expected projection error between the original and virtual cameras, enabling consistent 3D perception across diverse setups. Our framework serves as a plug-and-play module for existing 3D perception methods, improving their robustness to camera parameter variability. We validate our framework in CARLA by training and testing models on different camera configurations, demonstrating that our approach significantly reduces performance degradation while maintaining adaptability across diverse sensor setups. 
To summarize, we make the following key contributions in this paper:
\begin{itemize}[leftmargin=*]
    \item To the best of our knowledge, \textbf{UniDrive} presents the first comprehensive framework designed to generalize vision-centric 3D perception models across diverse camera configurations.
    \item We introduce a novel strategy that transforms images into a unified virtual camera space, enhancing robustness to camera parameter variations.
    \item We propose a virtual configuration optimization strategy that minimizes projection error, improving model generalization with minimal performance degradation.
    \item We contribute a systematic data generation platform along with a 160,000 frames multi-camera dataset, and benchmark evaluating perception models across varying camera configurations.
\end{itemize}


\vspace{-1mm}

\section{Related Work}

\vspace{-1mm}

\noindent\textbf{Vision-based 3D Detection.}
The development of camera-only 3D perception has gained great momentum recently. Early works such as FCOS3D~\citep{wang2021fcos3d}, which extended the 2D FCOS detector~\citep{tian2020fcos} by adding 3D object regression branches, paved the way for improvements in depth estimation via probabilistic modeling~\citep{wang2022probabilistic,chen2022epro}. 
Later methods like DETR3D~\citep{wang2022detr3d}, PETR~\citep{liu2022petr}, and Graph-DETR3D~\citep{chen2022graph} applied transformer-based architectures with learnable object queries in 3D space, drawing from the foundations of DETR~\citep{zhu2021deformable,wang2022anchor}, bypassing the limitations of perspective-based detection.
Recent works utilize bird's-eye view (BEV) for better 3D understanding. BEVDet~\citep{huang2021bevdet} and M²BEV~\citep{xie2204m2bev} effectively extended the Lift-Splat-Shoot (LSS) framework~\citep{philion2020lift} for 3D object detection. CaDDN~\citep{reading2021categorical} introduced explicit depth supervision in the BEV transformation to improve depth estimation. 
In addition, BEVFormer~\citep{li2022bevformer}, CVT~\citep{zhou2022cross}, and Ego3RT~\citep{lu2022learning} explored multi-head attention mechanisms for view transformation, demonstrating further improvements in consistency.
To further enhance accuracy, BEVDet4D~\citep{huang2022bevdet4d}, BEVFormer~\citep{li2022bevformer}, and PETRv2~\citep{liu2023petrv2} leveraged temporal cues in multi-camera object detection, showing significant improvements over single-frame methods.

\noindent\textbf{Cross Domain Perception.}
The cross-camera configurations problem proposed in this paper lies in the area of cross-domain perception. Domain generalization or adaptation is to enhance model performance on varying domains without re-training. For 2D perception, numerous cross-domain methods, such as feature distribution alignment and pseudo-labeling~\citep{muandet2013domain,li2018domain,dou2019domain,facil2019cam,chen2018domain,xu2020exploring,he2020domain,zhao2020collaborative}, have primarily addressed domain shifts caused by environmental factors like rain or low light. Recent 3D driving perception works~\citep{hao2024your,peng2023learning} focus on transfering the models trained on clean environment or perfect sensor situations to corrupted sensor and noisy environments, leading to several benchmarks and methods. Cross camera configuration is a relatively new topic in this area. While some works~\citep{wang2023towards} find that the model’s overfitting to camera parameters can lead to degrade performance because the models learn the fixed observation perspectives, the driving perception across camera parameters has seldom been systematically investigated. 

\noindent\textbf{Sensor Configuration.}
Sensor configurations has been proven important in the design of perception systems~\citep{joshi2008sensor,xu2022optimization}. Despite being relatively new in autonomous driving research~\citep{liu2019should}, sensor placement has gained significant attention. For instance, \citet{hu2022investigating} and \citet{li2024your} were the first to explore multi-LiDAR setups for improving 3D object detection, and \citet{li2024influence} studied how combining LiDAR and cameras impacts multi-modal detection systems. Several other studies \citep{jin2022roadside,kim2023placement,cai2023analyzing,jiang2023optimizing} focused on the strategic positioning of roadside LiDAR sensors for vehicle-to-everything (V2X) communication, shifting away from in-vehicle sensor placements. Although many efforts have aimed to refine sensor configurations for better performance, the challenge of adapting perception models to different sensor setups has been largely overlooked. Our research is the first to explore the generalization of driving perception models across diverse camera configurations.

\section{UniDrive}

\subsection{Problem Formulation}

In real-world multi-camera driving systems, perception models are typically trained on a specific camera configuration with fixed intrinsic and extrinsic parameters. However, the performance of these models often deteriorates when applied to new camera configurations, where the cameras may have different placements, orientations, or intrinsic properties. 


\textbf{Perception Across Multi-camera Configurations.} Given a set of cameras $\mathcal{C} =\{C_1, C_2, \dots, C_J\}$, each characterized by its intrinsic matrix $\mathbf{K}^{C_j} \in \mathbb{R}^{3 \times 3}$ and extrinsic matrix $\mathbf{E}^{C_j} \in \mathbb{R}^{4 \times 4}$, where $j \in \{1, 2, \dots, J\}$ and $J$ is the number of cameras. The images captured by these cameras are denoted as $\mathbf{I}^{C_j} \in \mathbb{R}^{H^{C_j} \times W^{C_j} \times 3}$, where $H^{C_j}$ and $W^{C_j}$ are the height and width of the image $\mathbf{I}^{C_j}$. When deploying the model trained on $\{C_1, C_2, \dots, C_J\}$ to a new set of cameras $\{C'_1, C'_2, \dots, C'_{J'}\}$ with different camera numbers and intrinsic and extrinsic parameters, the model may no longer effectively understand the 3D scene due to the differences between the training and testing configurations.

\textbf{Universal Multi-camera Representation.} To address the transferability of learned models across camera configurations, we attempt to design a universal representation, which transforms images from different camera configurations to a unified space before input to the deep learning network. 
To achieve this, we propose a \textbf{Virtual Camera Projection} approach, which re-projects the views $\mathbf{I}^{C_j}$ from the original cameras $\mathcal{C} =\{C_1, C_2, \dots, C_J\}$ into a unified set of virtual camera configurations $\mathcal{V} =\{V_1, V_2, ..., V_K\}$, where $K$ is the number of virtual cameras. The image is represented as $\mathbf{I}^{V_k} \in \mathbb{R}^{H^{V_k} \times W^{V_k} \times 3}$, where $H^{V_k}$ and $W^{V_K}$ are the image sizes, and $k \in \{1, 2, \dots, K\}$ indexes the virtual camera views. We denote $\mathbf{K}^{V_k}$ and $\mathbf{E}^{V_k}$ as the intrinsic and extrinsic matrices for the virtual camera $V_k$. This virtual configuration serves as a standardized coordinate system for both training and inference, allowing the model to operate consistently across different physical camera setups.

\begin{figure}[t]
    \centering
    \includegraphics[width=\textwidth]{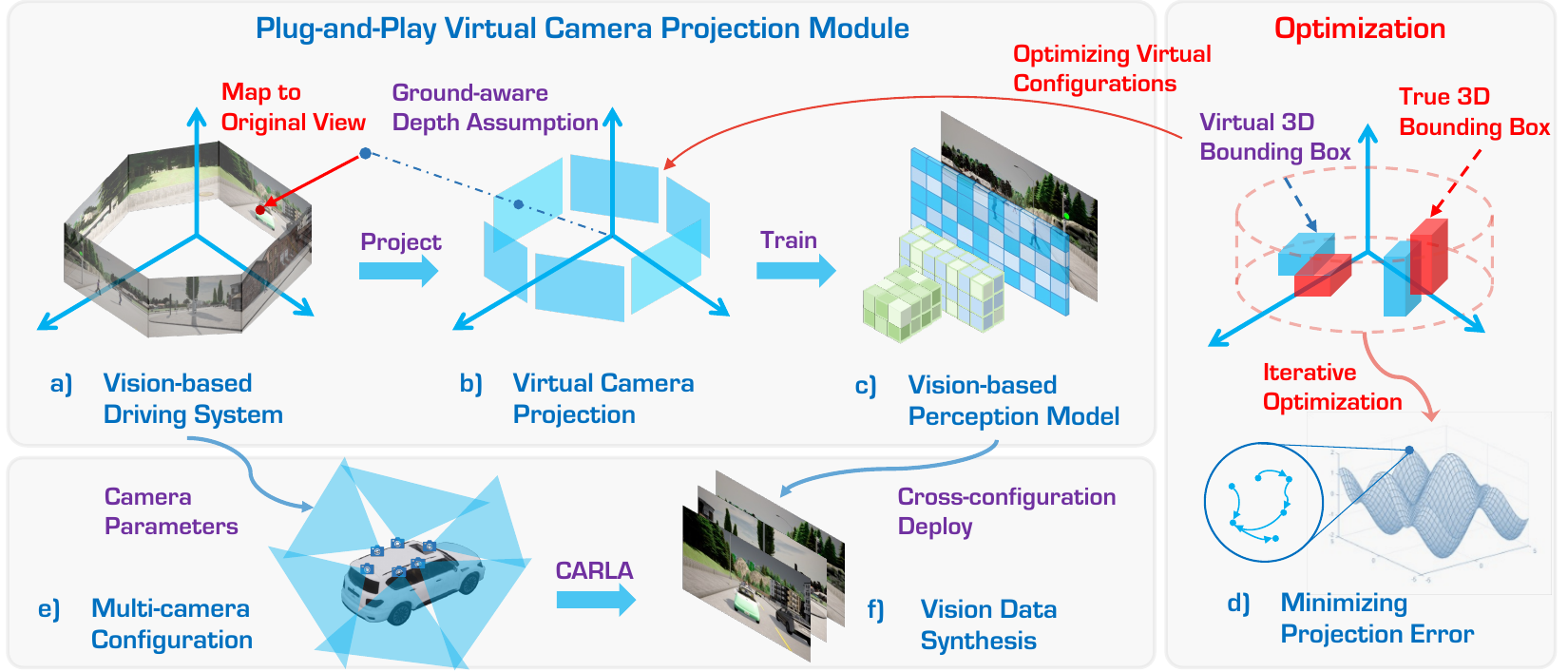}
    \caption{\textbf{Overview of UniDrive framework.} We transform the input images into a unified virtual camera space to achieve universal driving perception. To estimate the depth of pixels in the virtual view for projection, we propose a ground-aware depth assumption strategy. To obtain the most effective virtual camera space for multiple real camera configurations, we propose a data-driven CMA-ES~\citep{hansen2016cma} based optimization strategy. To evaluate the efficacy of our framework, we propose an automatic data generation platform in CARLA~\citep{Dosovitskiy17}.}
    \label{fig:overview}
    \vspace{-0.2cm}
\end{figure}

\subsection{Virtual Camera Projection}

In this subsection, we explain the Virtual Camera Projection method to project points from multiple camera views onto virtual camera views using a combination of ground and cylindrical surface assumptions, as shown in Figure~\ref{fig:overview}. The goal is to learn a transformation function $\mathcal{T}_{V \leftarrow C}$ that maps the images from the original cameras $\mathcal{C} = \{C_1, C_2, \dots, C_J\}$ to the virtual cameras $\mathcal{V} = \{V_1, V_2, \dots, V_K\}$ with minimum errors.

\textbf{Ground-aware Assumption.} For each pixel at coordinates $(u^{V_k}, v^{V_k})$ in the virtual view, its 3D coordinates in the virtual camera frame $(X^{V_k}_{\text{c}}, Y^{V_k}_{\text{c}}, Z^{V_k}_{\text{c}})$ are calculated based on the pixel's position in the image and the depth assumptions. Let the camera height be $h_{\text{c}}$, the focal lengths of the camera be $f_x^{V_k}$ and $f_y^{V_k}$, and the principal point (image center) be $(c_x^{V_k}, c_y^{V_k})$.
We first project all pixels to the ground plane to compute the initial assumption of 3D coordinates in virtual camera frame as,
\begin{align} \label{eq1}
\left(\hat{X}^{V_k}_{\text{c}},\, \hat{Y}^{V_k}_{\text{c}},\, \hat{Z}^{V_k}_{\text{c}}\right) = \left(\frac{f_y^{V_k} \, (u^{V_k} - c_x^{V_k})}{f_x^{V_k} \, (v^{V_k} - c_y^{V_k})}~ h_{\text{c}},\, h_{\text{c}},\, \frac{f_y^{V_k}}{v^{V_k} - c_y^{V_k}} ~h_{\text{c}} \right).
\end{align}
The Euclidean distance to optical center is computed as $\hat{D}_{\text{c}}^{V_k} = \left\|\begin{pmatrix} \hat{X}^{V_k}_{\text{c}}, \hat{Y}^{V_k}_{\text{c}}, \hat{Z}^{V_k}_{\text{c}} \end{pmatrix}\right\|_2$.
Then we compare the distance $\hat{D}_{\text{c}}^{V_k}$ with threshold $D_0$, if $\hat{D}_{\text{c}}^{V_k} < D_0$, the points connected to corresponding pixels in the images are assumed on the ground, $(X^{V_k}_{\text{c}}, Y^{V_k}_{\text{c}}, Z^{V_k}_{\text{c}}) = (\hat{X}^{V_k}_{\text{c}}, \hat{Y}^{V_k}_{\text{c}}, \hat{Z}^{V_k}_{\text{c}})$.
If $\hat{D}_{\text{c}}^{V_k} \ge D_0$, we assume that the points lie on a cylindrical-like surface at a fixed distance $D_0$ from the camera's optical center. In this case, the 3D coordinates are computed as:
\begin{align}\label{eq2}
\left(X^{V_k}_{\text{c}},\, Y^{V_k}_{\text{c}},\, Z^{V_k}_{\text{c}}\right) = \left(\frac{(u^{V_k} - c_x^{V_k})~ D_0}{f_x^{V_k}\,d^{V_k}},\, \frac{(v^{V_k} - c_y^{V_k}) ~ D_0}{f_y^{V_k}\,d^{V_k}},\, \frac{D_0}{d^{V_k}} \right),
\end{align}
where $d^{V_k} = \left\| \left( \frac{u^{V_k} - c_x^{V_k}}{f_x^{V_k}},\, \frac{v^{V_k} - c_y^{V_k}}{f_y^{V_k}},\, 1 \right) \right\|_2$.

\begin{wrapfigure}{R}{0.58\textwidth}

\vspace{-6mm}
\begin{minipage}{0.58\textwidth}

\begin{algorithm}[H]
    \fontsize{9}{10}\selectfont 
    \caption{Virtual Camera Projection}\label{alg:virtual_projection}
    \begin{algorithmic}[1]
    \State \textbf{Input:} $\{C_j, \mathbf{K}^{C_j}, \mathbf{E}^{C_j}, \mathbf{I}^{C_j}\}_{j=1}^{J}$, $\{V_k, \mathbf{K}^{V_k}, \mathbf{E}^{V_k}\}_{k=1}^{K}$
    \State \textbf{Output:} $\{\mathbf{I}^{V_k}\}_{k=1}^{K}$
    \For{$k = 1, 2, \dots, K$}
        \For{$(u^{V_k}, v^{V_k})$ in $\mathbf{I}^{V_k}$}
            \State Compute $(\hat{X}^{V_k}_{\text{c}}, \hat{Y}^{V_k}_{\text{c}}, \hat{Z}^{V_k}_{\text{c}})$, $\hat{D}_{\text{c}}^{V_k}$ using \eqref{eq1}
            \If{$\hat{D}_{\text{c}}^{V_k} < D_0$}
                \State $(X^{V_k}_{\text{c}}, Y^{V_k}_{\text{c}}, Z^{V_k}_{\text{c}}) \gets (\hat{X}^{V_k}_{\text{c}}, \hat{Y}^{V_k}_{\text{c}}, \hat{Z}^{V_k}_{\text{c}})$
            \Else
                \State Compute $(X^{V_k}_{\text{c}}, Y^{V_k}_{\text{c}}, Z^{V_k}_{\text{c}})$ using \eqref{eq2}
            \EndIf
            \State $\mathbf{p}_{\text{w}} \gets \mathbf{E}^{V_k} \cdot \mathbf{p}^{V_k}_{\text{c}}$, $\mathbf{p}^{V_k}_{\text{c}} = (X^{V_k}_{\text{c}}, Y^{V_k}_{\text{c}}, Z^{V_k}_{\text{c}})$
            \State $\mathbf{p}^{C_j}_{\text{c}} \gets {\mathbf{E}^{C_j}}^{-1} \cdot \mathbf{p}_{\text{w}}$
            \State $(u^{C_j}, v^{C_j}) \gets \mathbf{K}^{C_j} \cdot \mathbf{p}^{C_j}_{\text{c}}$
            \State $\mathbf{I}^{V_k \leftarrow C_j}(u^{V_k}, v^{V_k}) \gets \mathbf{I}^{C_j}(u^{C_j}, v^{C_j})$
        \EndFor
    \EndFor
    \State $\mathbf{I}^{V_k} \gets \frac{1}{\mathbf{W}} \sum_{j=1}^{J} w_j \cdot \mathbf{I}^{V_k \leftarrow C_j}$ using \eqref{eq3}
    \end{algorithmic}
    \label{alg1}
\end{algorithm}

\vspace{-8mm}
\end{minipage}
\end{wrapfigure}

\textbf{Point-wise Projection.}
Once the 3D coordinates $(X^{V_k}_{\text{c}}, Y^{V_k}_{\text{c}}, Z^{V_k}_{\text{c}})$ in the virtual camera frame are calculated, we transform the point into the world coordinate system with extrinsic matrix $\mathbf{E}^{V_k}$, $\mathbf{p}_{\text{w}} = \mathbf{E}^{V_k} \cdot \mathbf{p}^{V_k}_{\text{c}}$,
where $\mathbf{p}^{V_k}_{\text{c}} = (X^{V_k}_{\text{c}}, Y^{V_k}_{\text{c}}, Z^{V_k}_{\text{c}}, 1)^\top$ is the homogeneous coordinate of the point in the virtual camera's frame, and $\mathbf{p}_{\text{w}} \in \mathbb{R}^4$ is the 3D point in the world coordinate system.
Next, we transform the point from the world coordinate system into the original camera's coordinate system using the inverse of the original camera's extrinsic matrix $\mathbf{p}^{C_j}_{\text{c}} = {\mathbf{E}^{C_j}}^{-1} \cdot \mathbf{p}_{\text{w}}$.
Finally, we project the point back onto the original camera's 2D image plane using its intrinsic matrix, $(u^{C_j}, v^{C_j}, 1)^\top = \mathbf{K}^{C_j} \cdot \mathbf{p}^{C_j}_c = \mathbf{K}^{C_j} \cdot {\mathbf{E}^{C_j}}^{-1} \cdot \mathbf{p}_{\text{w}}$.
This provides the pixel coordinates $(u^{C_j}, v^{C_j})$ in the original view that correspond to the pixel $(u^{V_k}, v^{V_k})$ in the virtual view. We denote $\mathbf{P}_{V_k \leftarrow C_j}(\hat{D}_{\text{c}}^{V_k})$ as the projection transform matrix from $(u^{C_j}, v^{C_j})$ in the $i$-th original view to $(u^{V_k}, v^{V_k})$ based on the Euclidean distance to virtual camera optical center $\hat{D}_{\text{c}}^{V_k}$.

\textbf{Image-level Transformation.}
The point-wise projection is extended to the entire image view. For each pixel $(u^{V_k}, v^{V_k})$ in the $k$-th virtual view, we compute the corresponding pixel $(u^{C_j}, v^{C_j})$ in the $i$-th original view based on the projection matrix $\mathbf{P}_{V_k \leftarrow C_j}(\hat{D}_{\text{c}}^{V_k})$. The entire image $\mathbf{I}^{C_j}$ of the $i$-th original view is warped into the virtual view $\mathbf{I}^{V_k \leftarrow C_j}$ as follows, $\mathbf{I}^{V_k \leftarrow C_j} = \mathcal{T}(\mathbf{I}^{C_j}, \mathbf{P}_{V_k \leftarrow C_j}(\hat{D}_{\text{c}}^{V_k}))$, where $\mathcal{T}(\mathbf{I}, \mathbf{P})$ represents the warping function applied to the image $\mathbf{I}^{C_j}$ using the projection matrix $\mathbf{P}_{V_k \leftarrow C_j}(\hat{D}_{\text{c}}^{V_k})$ based on the $\hat{D}_{\text{c}}^{V_k}$.

\textbf{Blending Multiple Views.}
Since each pixel in a single virtual view may have corresponding pixels from various original view, after transforming each original view into the virtual view, we merge all the transformed images $\mathbf{I}^{V_k \leftarrow C_j}$ to form the final output image $\mathbf{I}^{V_k}$. This blending is performed by computing a weighted sum of all the projected views:
\begin{align}\label{eq3}
\mathbf{I}^{V_k} = \frac{1}{\mathbf{W}} \sum_{j=1}^{J} w_j \cdot \mathbf{I}^{V_k \leftarrow C_j} = \frac{1}{\mathbf{W}} \sum_{i=1}^J w_j \cdot \mathcal{T} (\mathbf{I}^{C_j}, \mathbf{P}_{V_k \leftarrow C_j}(\hat{D}_{\text{c}}^{V_k})),
\end{align}
where $\mathbf{W} = \sum_{j=1}^{J} w_j$ is the total weight, and $w_j$ is the blending weight for the $j$-th original view. The weights can be based on factors such as the angular distance between the original and virtual views, or the proximity of the cameras.
We presented the detailed computation process in Algorithm~\ref{alg1}.

\subsection{Virtual Projection Error}

To evaluate the accuracy of the Virtual Camera Projection method in the context of a 3D object detection task, we propose a weighted projection error metric based on angular discrepancies between the virtual and original camera views. This method accounts for both angular deviations and the distance from the camera’s optical center to provide a more robust error evaluation.

\textbf{Angle Computation.}
Given a driving scenario of 3D bounding box information, for each 3D bounding box $b_n = \{(x_{n,m}, y_{n,m}, z_{n,m})^\top \}_{m=1}^8$, we first project its corner points onto the original camera $C_j$ as the pixel $(u^{C_j}_{n,m}, v^{C_j}_{n,m})$, using the intrinsic matrix $\mathbf{K}^{C_j}$ and extrinsic matrix $\mathbf{E}^{C_j}$. 
Then, we use the inverse of the warping process $\mathbf{P}_{V_k \leftarrow C_j}$ to find the corresponding pixel $(u^{V_k}_{n,m}, v^{V_k}_{n,m})$ in the virtual camera view $V_k$ for each corner point.
We compute the pitch angle $\theta^{V_k}_{n,m}$ and yaw angle $\phi^{V_k}_{n,m}$ relative to the virtual camera's optical center:
\begin{align} 
\left(\theta^{V_k}_{n,m} ,\, \phi^{V_k}_{n,m} \right) = \left( \arctan \frac{v^{V_k}_{n,m} - c_y^{V_k}}{f_y^{V_k}} ,\, \arctan \frac{u^{V_k}_{n,m} - c_x^{V_k}}{f_x^{V_k}} \right).
\end{align} 
Next, for the same corner points, we directly project to the virtual view using $\mathbf{K}^{V_k}$ and $\mathbf{E}^{V_k}$ as $(u^{V_k\prime}_{n,m}, v^{V_k\prime}_{n,m})$. Then the pitch angle $\theta^{V_k\prime}_{n,m}$ and yaw angle $\phi^{V_k\prime}_{n,m}$ are
\begin{align}
\left( \theta^{V_k\prime}_{n,m},\, \phi^{V_k\prime}_{n,m} \right) = \left( \arctan \frac{v^{V_k\prime}_{n,m} - c_y^{V_k}}{f_y^{V_k}} ,\, \arctan \frac{u^{V_k\prime}_{n,m} - c_x^{V_k}}{f_x^{V_k}} \right).
\end{align}
\textbf{Angle Error Calculation.} For each corner point, we compute the angular error between the original camera projection and the corresponding point in the virtual camera. The absolute errors in pitch and yaw are
$ \Delta \theta_{n,m}^{V_k} = \left| \theta^{V_k}_{n,m} - \theta^{V_k\prime}_{n,m} \right|, $
$ \Delta \phi_{n,m}^{V_k} = \left| \phi^{V_k}_{n,m} - \phi^{V_k\prime}_{n,m} \right|. $
We use the distance $D_{n,m}^{V_k}$ of each corner point from the original camera's optical center as a weight. The distance is computed as
$ D_{n,m}^{V_k} = \left\|\begin{pmatrix} x^{V_k}_{n,m}, y^{V_k}_{n,m}, z^{V_k}_{n,m} \end{pmatrix}\right\|$.
The weighted error for each corner point is then calculated as
$ \mathcal{E}_{n,m}^{V_k} = D_{n,m}^{V_k} \cdot ( \Delta \theta_{n,m}^{V_k} + \Delta \phi_{n,m}^{V_k} ). $
The overall error for a 3D bounding box $b_n$ is obtained by summing the weighted errors of its eight corner points
$ \mathcal{E}_{b_n}^{V_k} = \sum_{m=1}^8 \mathcal{E}_{n,m}^{V_k}. $
We sum the projection errors across all 3D bounding boxes $b_n \in \mathcal{B}$ to compute the total projection error \begin{align}\mathcal{E} = \sum_{n=1}^{N} \mathcal{E}_{b_n}^{V_k} = \sum_{n=1}^{N} \sum_{m=1}^8 \mathcal{E}_{n,m}^{V_k}. \end{align}

\begin{figure}[t]
    \centering
    \includegraphics[width=\textwidth]{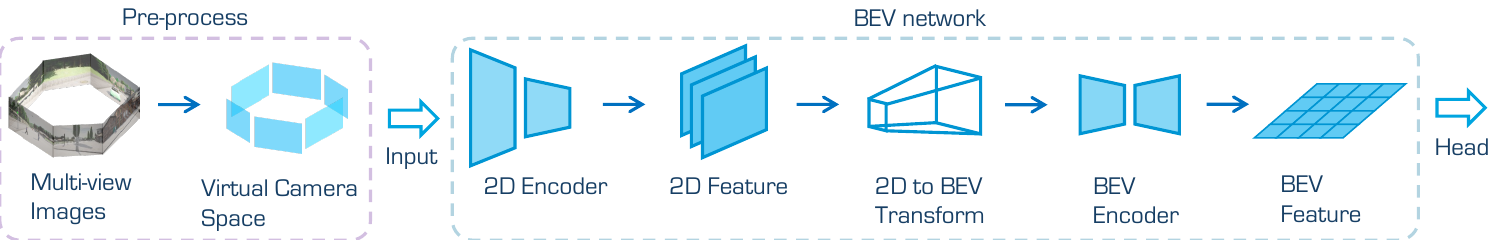}
    \caption{\textbf{Integration with Existing Methods.} Our virtual camera projection can be integrated into the pipeline as a pre-processing stage before feeding the multi-view images into the network.}
    \label{fig:bev_pipeline}
\end{figure}

\begin{algorithm}[t]
    \fontsize{9}{10}\selectfont 
    \caption{Virtual Camera Configuration Optimization}\label{alg:cmaes_virtual}
    \begin{algorithmic}[1]
    \State Initialize: $t \gets 0$, $\mathbf{m}^{(0)}$, $\sigma^{(0)}$, $\mathbf{C}^{(0)}$, $N_t$, $M_t$, $\forall t \in \{0, 1,2,\dots,T\}$
    \For{$t = 0, 1,2,\ldots, T$}
        \For{$i = 1$ to $N_t$}
            \State Sample $\mathbf{u}_i^{(t)} \sim \mathcal{N}(\mathbf{m}^{(t)}, (\sigma^{(t)})^2 \mathbf{C}^{(t)})$ from $\delta$-density gird-level  candidates
            \State Calculate $\mathcal{E}(\mathbf{u}_i^{(t)})$
        \EndFor
        \State Update $\mathbf{m}^{(t+1)}$ based on the top $M_t$ best solutions $\hat{\mathbf{u}}_{i}^{(t)}$ via  \eqref{eq:m_new}
        \State Update $\sigma^{(t+1)}$ and $\mathbf{C}^{(t+1)}$  via \eqref{eq:p_c_new}, \eqref{eq:C_new}, \eqref{eq:p_sigma_new}, and \eqref{eq:sigma_new}\EndFor
    \end{algorithmic}
    \label{alg2}
\end{algorithm}

\subsection{Optimizing Virtual Camera Configurations}

Given a set of multi-camera systems, we aim to design a unified virtual camera configuration that minimizes the reprojection error across all original camera configurations. To achieve this, we adopt the heuristic optimization based on the Covariance Matrix Adaptation Evolution Strategy (CMA-ES) \citep{hansen2016cma} to find an optimized set of virtual camera configurations. 

\textbf{Objective Function.} Given multiple driving perception systems with varying multi-camera confgirations indexed by $s$, the total error across all systems is expressed as $\mathcal{E}_{total} = \sum_{s=1}^{S} \mathcal{E}^{(s)}(\mathbf{u})$, where $\mathbf{u} = \{V_k, \mathbf{K}^{V_k}, \mathbf{E}^{V_k}\}_{k=1}^{K}$ includes both the intrinsic and extrinsic camera parameters of virtual multi-camera framework, $K$ is the total quantity of virtual cameras and $S$ is the total quantity of multi-camera driving systems that share the same perception model. We aim to minimize this error by sampling and updating the virtual camera parameters iteratively through a CMA-ES based optimization method. 


\noindent\textbf{Optimization Method.} 
Our Optimization strategy begins by defining a multivariate normal distribution $\mathcal{N}(\mathbf{m}^{(t)}, (\sigma^{(t)})^2 \mathbf{C}^{(t)})$, where $\mathbf{m}^{(t)}$ represents the mean vector, $\sigma^{(t)}$ denotes the step size, and $\mathbf{C}^{(t)}$ is the covariance matrix at iteration $t$. The configuration space $\mathcal{U}$ is discretized with a density $\delta$, and $N_t$ candidate configurations $\mathbf{u}_i^{(t)}\sim\mathcal{N}(\mathbf{m}^{(t)}, (\sigma^{(t)})^2 \mathbf{C}^{(t)})$ are sampled at each iteration $t$. Initialization begins with the initial mean $\mathbf{m}^{(0)}$, step size $\sigma^{(0)}$, and covariance matrix $\mathbf{C}^{(0)} = \mathbf{I}$. The updated mean vector $\mathbf{m}^{(t+1)}$ is calculated in the subsequent iteration to serve as the new center for the search distribution concerning the virtual camera configuration. The process can be mathematically expressed as:
\begin{equation}
\label{eq:m_new}
\mathbf{m}^{(t+1)} = \sum_{i=1}^{M_t} w_i \hat{\mathbf{u}}_{i}^{(t)}, \; \mathcal{E}(\hat{\mathbf{u}}_1^{(t)}) \geq \mathcal{E}(\hat{\mathbf{u}}_2^{(t)}) \geq \dots \geq \mathcal{E}(\hat{\mathbf{u}}_{M_t}^{(t)})~,
\end{equation}
where $M_t$ is the number of top solutions selected to update $\mathbf{m}^{(t+1)}$, and $w_i$ are weights determined by solution performance. The evolution path $\mathbf{p}_{\mathbf{C}}^{(t+1)}$, which tracks the direction of successful optimization steps, is updated as:
\begin{align}
\label{eq:p_c_new}
    \mathbf{p}_{\mathbf{C}}^{(t+1)} =  (1 - c_{\mathbf{C}})\cdot \mathbf{p}_{\mathbf{C}}^{(t)} +\sqrt{1 - (1 - c_{\mathbf{C}})^2} \cdot \sqrt{\frac{1}{\sum_{i=1}^{M_t} w_i^2}} \cdot \frac{\mathbf{m}^{(t+1)} - \mathbf{m}^{(t)}}{\sigma^{(t)}}~,
\end{align}
where $c_{\mathbf{C}}$ is the learning rate for updating the covariance matrix. The covariance matrix $\mathbf{C}$, which defines the distribution's shape for camera configurations, is adjusted at each iteration as follows:
\begin{equation}\label{eq:C_new}
    \mathbf{C}^{(t+1)} = (1 - c_{\mathbf{C}}) \mathbf{C}^{(t)} + c_{\mathbf{C}} \mathbf{p}_{\mathbf{C}}^{(t+1)} {\mathbf{p}_{\mathbf{C}}^{(t+1)}}^T~.
\end{equation}
Similarly, the evolution path for the step size, $\mathbf{p}_{\sigma}$, is updated, and the global step size $\sigma$ is then adjusted to balance exploration and exploitation:
\begin{equation}
\label{eq:p_sigma_new}
\mathbf{p}_{\sigma}^{(t+1)} = (1 - c_{\sigma})\mathbf{p}_{\sigma}^{(t)} + \sqrt{1 - (1 - c_{\sigma})^2} \cdot \sqrt{\frac{1}{\sum_{i=1}^{M_t} w_i^2}} \cdot \frac{\mathbf{m}^{(t+1)} - \mathbf{m}^{(t)}}{\sigma^{(t)}}~,
\end{equation}
\begin{equation}
\label{eq:sigma_new}
\sigma^{(t+1)} = \sigma^{(t)} \exp\left(\frac{c_{\sigma}}{d_{\sigma}} \left(\frac{\|\mathbf{p}_{\sigma}^{(t+1)}\|}{E\|\mathcal{N}(0, \mathbf{I})\|} - 1\right)\right)~,
\end{equation}
where $c_{\sigma}$ is the learning rate for updating $\mathbf{p}_{\sigma}$, and $d_\sigma$ is a normalization factor controlling the adjustment rate of the global step size.
We presented the detailed optimization process in Algorithm~\ref{alg2}.

\begin{figure}[t]
    \centering
    \begin{subfigure}[h]{0.19\textwidth}
        \centering
        \includegraphics[width=\textwidth]{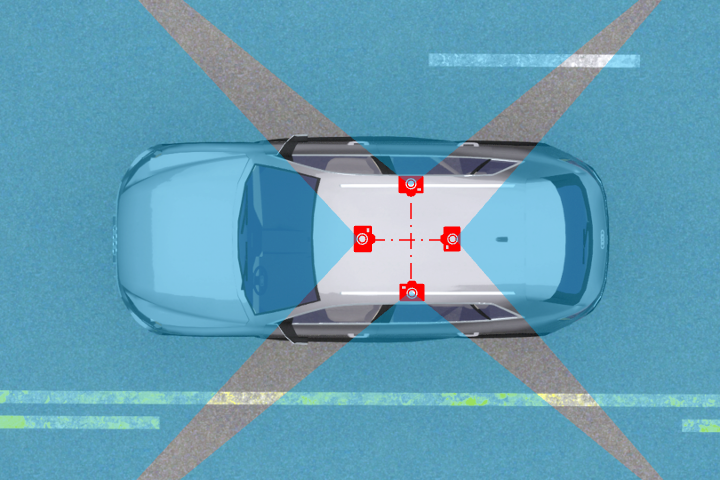}
        \caption{$4\times95^\circ$}
        \label{fig:4x}
    \end{subfigure}
    \begin{subfigure}[h]{0.19\textwidth}
        \centering
        \includegraphics[width=\textwidth]{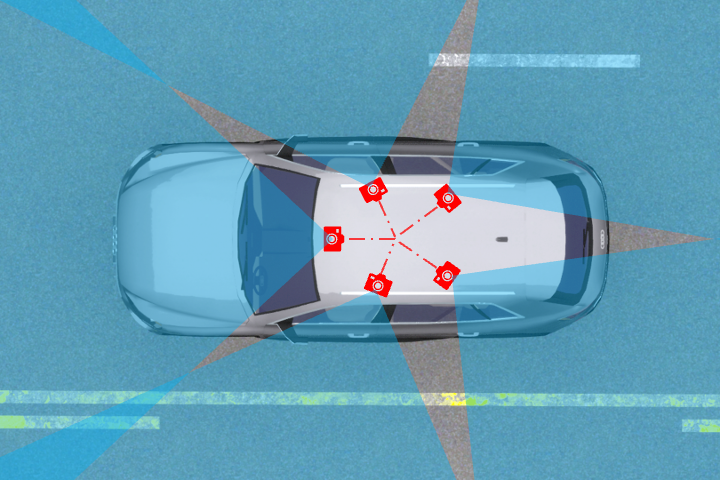}
        \caption{$5\times75^\circ$}
        \label{fig:5x}
    \end{subfigure}
    \begin{subfigure}[h]{0.19\textwidth}
        \centering
        \includegraphics[width=\textwidth]{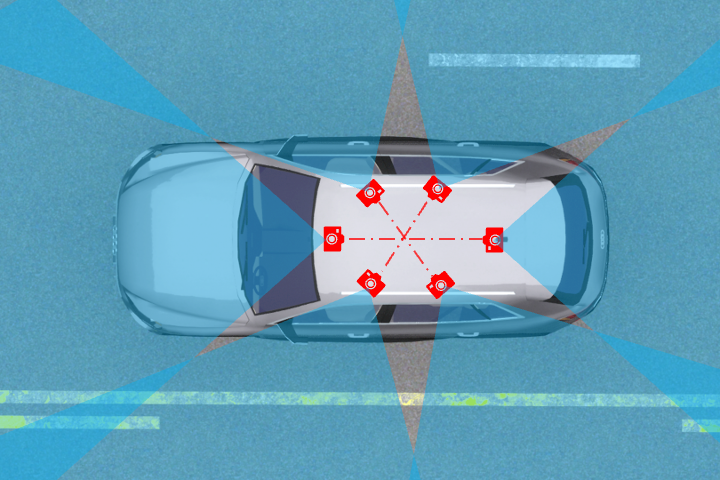}
        \caption{$6\times80^\circ a$}
        \label{fig:6x80I}
    \end{subfigure}
    \begin{subfigure}[h]{0.19\textwidth}
        \centering
        \includegraphics[width=\textwidth]{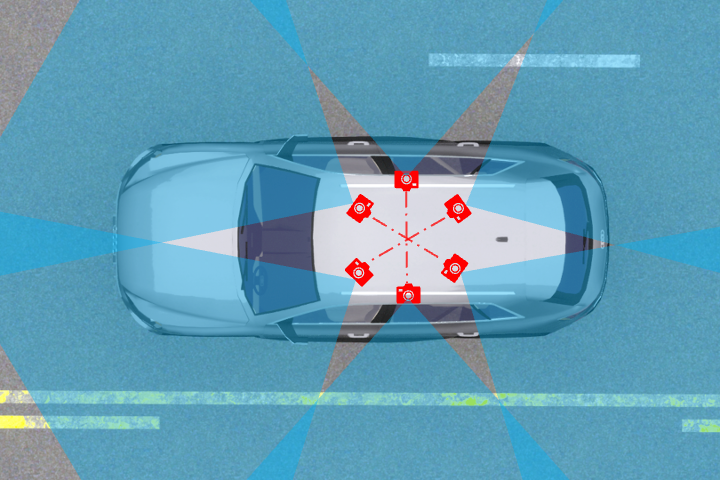}
        \caption{$6\times80^\circ b$}
        \label{fig:6x80II}
    \end{subfigure}
    \begin{subfigure}[h]{0.19\textwidth}
        \centering
        \includegraphics[width=\textwidth]{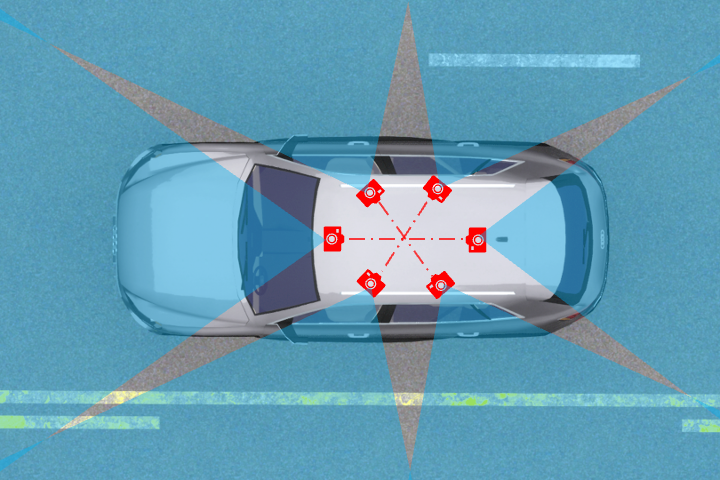}
        \caption{$6\times70^\circ$}
        \label{fig:6x70}
    \end{subfigure}
    \begin{subfigure}[h]{0.19\textwidth}
        \centering
        \includegraphics[width=\textwidth]{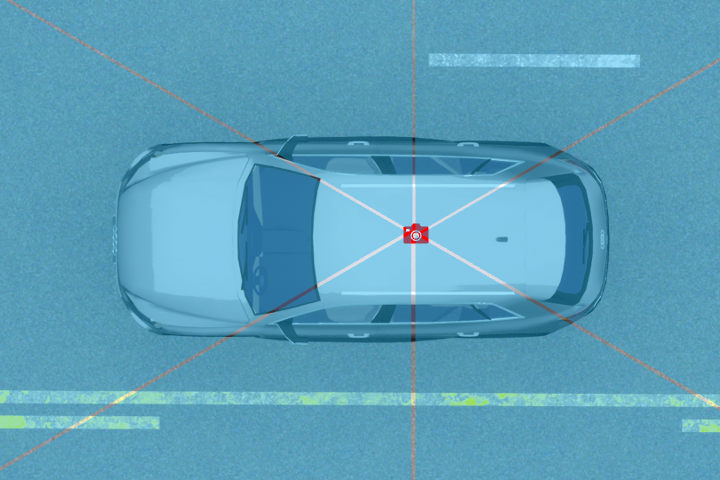}
        \caption{$6\times60^\circ$}
        \label{fig:6x60}
    \end{subfigure}
    \begin{subfigure}[h]{0.19\textwidth}
        \centering
        \includegraphics[width=\textwidth]{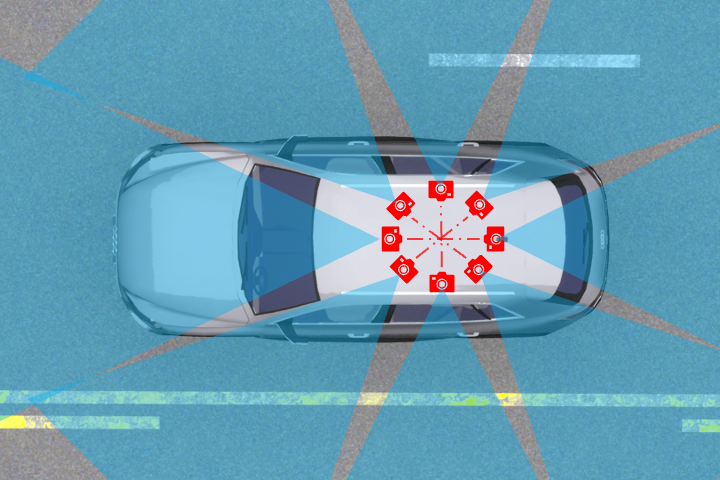}
        \caption{$8\times50^\circ$}
        \label{fig:8x}
    \end{subfigure}
    \begin{subfigure}[h]{0.19\textwidth}
        \centering
        \includegraphics[width=\textwidth]{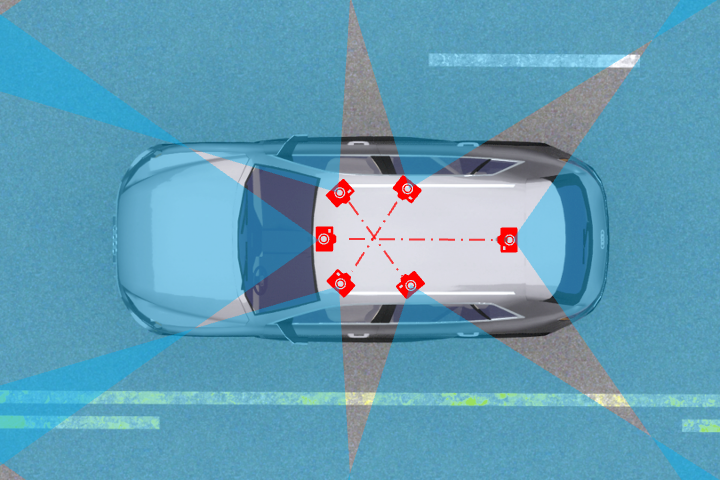}
        \caption{$5\times70^\circ + 110^\circ$}
        \label{fig:nusc}
    \end{subfigure}
    \begin{subfigure}[h]{0.19\textwidth}
        \centering
        \includegraphics[width=\textwidth]{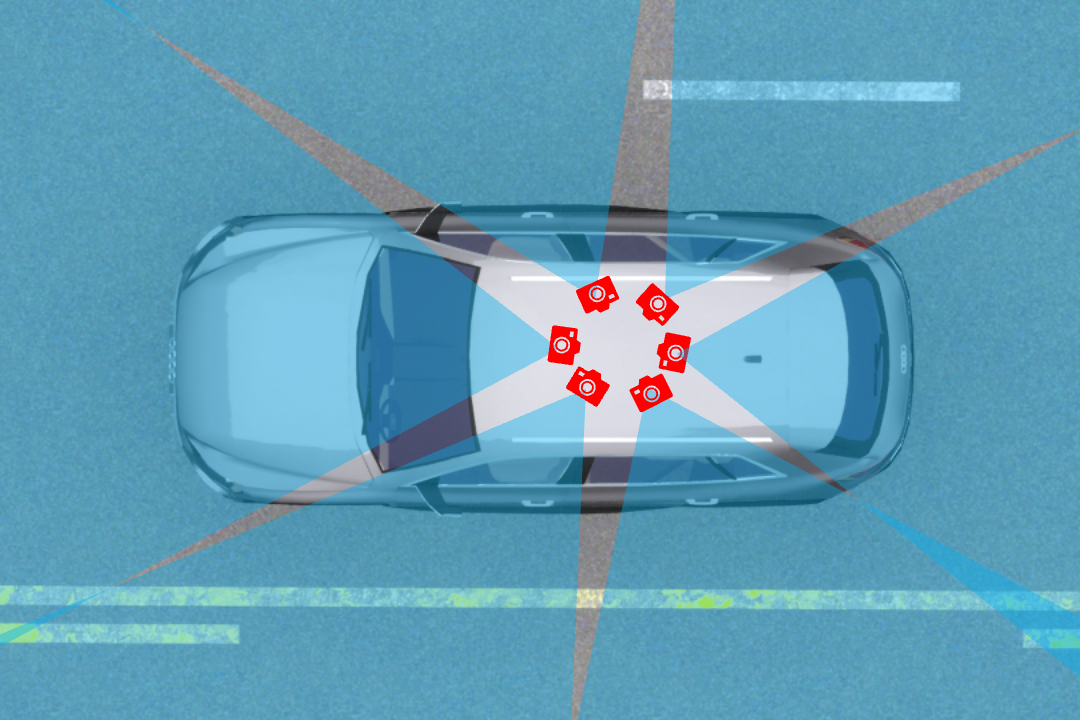}
        \caption{w/o optimization}
        \label{fig:woo}
    \end{subfigure}
    \begin{subfigure}[h]{0.19\textwidth}
        \centering
        \includegraphics[width=\textwidth]{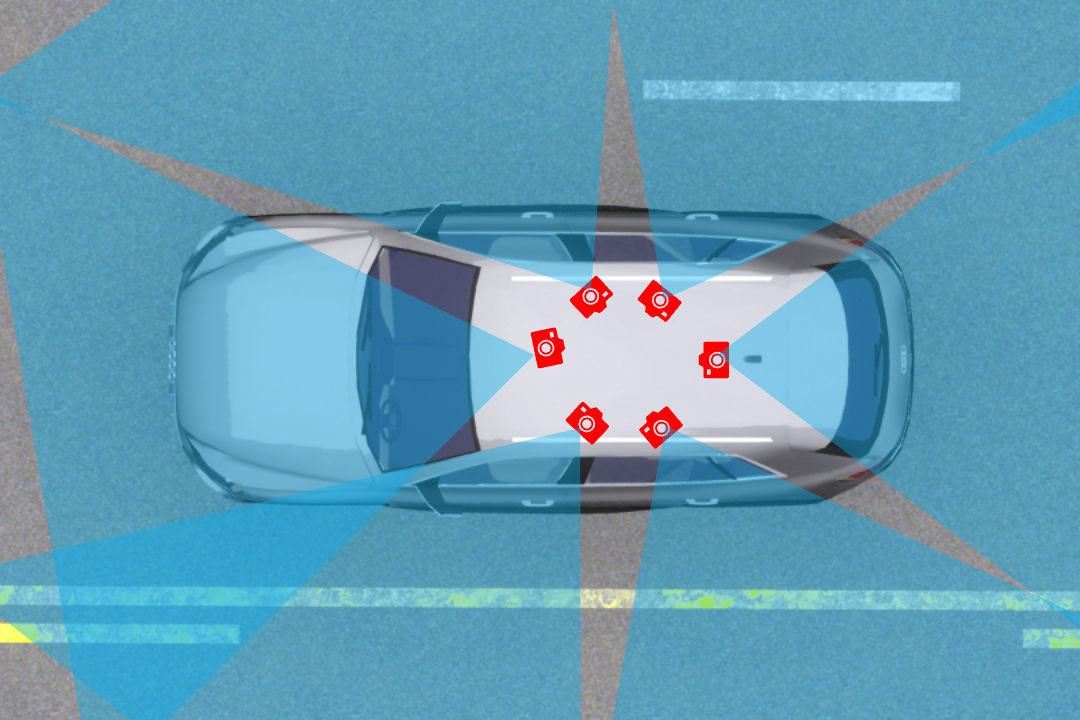}
        \caption{w/ optimization}
        \label{fig:wo}
    \end{subfigure}
    \caption{\textbf{Visualized multi-camera configurations.} We illustrate the multi-view camera configurations used in our study. There configurations are inspired by practical applications in the industry.}
    \vspace{-0.2cm}
    \label{fig:camera_configs}
\end{figure}

\section{Experiments}

\subsection{Benchmark Setups}

\textbf{Data Generation.} We generate multi-view image data and 3D objects ground truth in CARLA simulator~\citep{Dosovitskiy17}. We use the maps of Towns 1-6 to collect data. We incorporate 6 classes for 3D object detection, including \textit{Car}, \textit{Bus}, \textit{Truck}, \textit{Motorcycle}, \textit{Bicycle}, and \textit{Pedestrian}. The dataset consists of 500 scenes (20,000 frames) for each camera configuration. We split 250 scenes for training and 250 scenes for validation. Our dataset is organized as the format of nuScenes~\citep{nuScenes} and compatible to the {\tt nuscenes-devkit} python package for convenient processing.

\textbf{Camera Configurations.} We adopt several commonly used camera configurations in automotive practice with various camera quantities, placements and field of views. These configurations are represented in Figure~\ref{fig:camera_configs}. We set all camera resolutions to {\tt 1600$\times$900} as nuScenes. Our camera configurations include camera numbers from 4 to 8. For the field of view (FOV) for cameras, we conduct study mainly on 6 cameras with FOV = 60, 70, 80. For the placement, we design two different types of placement as shown in Figure~\ref{fig:camera_configs} (c), (d). We also include the original configurations of nuScenes~\citep{nuScenes} dataset with five $70^\circ$ cameras and a $110^\circ$ camera.

\textbf{Deteciton Method.} Due to the extensive computation resource needed to benchmark the multi-camera configurations, we only compare our method with the camera variant of BEVFusion~\citep{BEVFusion} (abbreviated as BEVFusion-C). 
BEVFusion is one of the representative methods in many leaderboards, such as nuScenes~\citep{nuScenes} and Waymo~\citep{sun2020waymoOpen}. 

\begin{table}[t]
\centering
\caption{\textbf{Quantitative results of BEVFusion-C for 3D detection across camera configurations}. The detector is trained on the \colorbox{unidrive_blue!8}{\textcolor{unidrive_blue}{blue}} configurations and tested on all configurations directly. We report the mAP ($\uparrow$) and class-level AP ($\uparrow$) scores in percentage ($\%$). }
\label{tab:bevfusion}
\renewcommand\arraystretch{1.2}
\setlength\tabcolsep{1pt}
\resizebox{\textwidth}{!}{
\begin{tabular}{c|p{0.90cm}<{\centering}|p{0.90cm}<{\centering}p{0.90cm}<{\centering}p{0.90cm}<{\centering}p{0.90cm}<{\centering}p{0.90cm}<{\centering}p{0.90cm}<{\centering}|c|p{0.90cm}<{\centering}|p{0.90cm}<{\centering}p{0.90cm}<{\centering}p{0.90cm}<{\centering}p{0.90cm}<{\centering}p{0.90cm}<{\centering}p{0.90cm}<{\centering}}
\toprule
\textbf{Configurations} & \textbf{mAP} & \textit{Car} & \textit{Bus} & \textit{Truck} & \textit{Ped.} & \textit{Motor} & \textit{Bic.} & \textbf{Configurations} & \textbf{mAP} & \textit{Car} & \textit{Bus} & \textit{Truck} & \textit{Ped.} & \textit{Motor} & \textit{Bic.} 
\\
\midrule\midrule
\rowcolor{unidrive_blue!8} \textcolor{unidrive_blue}{$5\times70^\circ + 110^\circ$} & \textcolor{unidrive_blue}{63.9} & \textcolor{unidrive_blue}{62.4} & \textcolor{unidrive_blue}{58.0} & \textcolor{unidrive_blue}{66.5} & \textcolor{unidrive_blue}{54.7} & \textcolor{unidrive_blue}{68.7} & \textcolor{unidrive_blue}{72.9} & \textcolor{unidrive_blue}{$6 \times 60$} & \textcolor{unidrive_blue}{69.3} & \textcolor{unidrive_blue}{68.7} & \textcolor{unidrive_blue}{67.4} & \textcolor{unidrive_blue}{66.3} & \textcolor{unidrive_blue}{62.2} & \textcolor{unidrive_blue}{78.4} & \textcolor{unidrive_blue}{72.8}
\\
\midrule
$4 \times 95^\circ$ & 4.9 & 4.6 & 5.1 & 3.8 & 3.9 & 3.1 & 4.2 & $4 \times 95^\circ$ & 3.8 & 4.1 & 4.3 & 3.2 & 4.1 & 3.3 & 3.6
\\
$5 \times 75^\circ$ & 7.2 & 9.5 & 4.8 & 6.2 & 5.1 & 9.0 & 8.3 & $5 \times 75^\circ$ & 3.4 & 3.7 & 2.5 & 3.1 & 2.7 & 4.3 & 4.2
\\
$6 \times 80^\circ a$ & 8.5 & 11.7 & 8.8 & 8.4 & 6.1 & 8.2 & 7.7 & $6 \times 80^\circ a$ & 0.6 & 1.7 & 0.4 & 0.5 & 0.1 & 0.4 & 0.6
\\
$6 \times 80^\circ b$ & 6.9 & 10.0 & 7.2 & 7.8 & 5.2 & 6.1 & 5.6 & $6 \times 80^\circ b$ & 0.4 & 1.4 & 0.0 & 0.7 & 0.0 & 0.2 & 0.1
\\
$6 \times 70^\circ$ & 67.5 & 65.2 & 61.2 & 69.3 & 57.9 & 79.5 & 72.1 & $6 \times 70^\circ$ & 8.1 & 9.6 & 4.3 & 6.8 & 7.1 & 11.0 & 10.0
\\
$6 \times 60^\circ$ & 9.2 & 12.4 & 7.0 & 8.0 & 6.5 & 11.9 & 9.4 & $5\times70^\circ + 110^\circ$ & 4.6 & 4.9 & 3.0 & 3.5 & 3.4 & 5.4 & 7.4
\\
$8 \times 50^\circ$ & 0.5 & 0.6 & 0.1 & 0.9 & 0.2 & 0.3 & 0.6 & $8 \times 50^\circ$ & 17.3 & 18.5 & 9.9 & 14.1 & 16.7 & 21.2 & 23.4
\\
\midrule\midrule
\rowcolor{unidrive_blue!8} \textcolor{unidrive_blue}{$6 \times 80^\circ a$} & \textcolor{unidrive_blue}{66.7} & \textcolor{unidrive_blue}{65.4} & \textcolor{unidrive_blue}{66.2} & \textcolor{unidrive_blue}{63.7} & \textcolor{unidrive_blue}{55.8} & \textcolor{unidrive_blue}{75.9} & \textcolor{unidrive_blue}{72.9} & \textcolor{unidrive_blue}{$6 \times 80^\circ b$} & \textcolor{unidrive_blue}{69.1} & \textcolor{unidrive_blue}{66.0} & \textcolor{unidrive_blue}{65.1} & \textcolor{unidrive_blue}{72.1} & \textcolor{unidrive_blue}{58.3} & \textcolor{unidrive_blue}{78.6} & \textcolor{unidrive_blue}{74.2}
\\
\midrule
$4 \times 95^\circ$ & 3.8 & 4.3 & 5.0 & 3.6 & 3.2 & 2.8 & 3.9 & $4 \times 95^\circ$ & 3.5 & 3.9 & 4.1 & 3.3 & 3.2 & 2.6 & 3.7 
\\
$5 \times 75^\circ$ & 30.4 & 31.2 & 23.4 & 27.8 & 28.6 & 36.9 & 34.2 & $5 \times 75^\circ$ & 29.6 & 30.3 & 22.6 & 27.1 & 27.9 & 36.3 & 33.2 
\\
$5\times70^\circ + 110^\circ$ & 9.2 & 10.5 & 6.5 & 8.6 & 7.1 & 8.8 & 13.3 & $6 \times 80^\circ a$ & 63.2 & 65.5 & 67.0 & 66.4 & 46.7 & 68.2 & 65.1
\\
$6 \times 80^\circ b$ & 63.3 & 65.4 & 63.8 & 70.9 & 46.3 & 68.1 & 65.4 & $6 \times 60^\circ$ & 1.7 & 2.9 & 0.5 & 1.1 & 0.7 & 2.3 & 2.6
\\
$6 \times 70^\circ$ & 16.4 & 18.0 & 9.4 & 13.3 & 14.7 & 22.2 & 20.6 & $6 \times 70^\circ$ & 16.1 & 17.7 & 8.3 & 12.3 & 14.6 & 23.7 & 19.9
\\
$6 \times 60^\circ$ & 1.8 & 3.3 & 0.8 & 1.5 & 0.6 & 2.3 & 2.4 & $5\times70^\circ + 110^\circ$ & 8.9 & 10.3 & 5.6 & 7.5 & 7.1 & 9.6 & 13.4
\\
$8 \times 50^\circ$ & 0.4 & 0.5 & 0.0 & 0.8 & 0.1 & 0.1 & 0.6 & $8 \times 50^\circ$ & 0.2 & 0.4 & 0.0 & 0.9 & 0.3 & 0.3 & 0.4 
\\
\bottomrule
\end{tabular}
}
\vspace{-2mm}
\end{table}

\subsection{Comparative Study}

We conduct comparative studies to evaluate the performance of camera perception across configurations. Through our analysis, we are able to demonstrate the effectiveness of UniDrive framework.

\begin{figure}[t]
    \centering
    \begin{subfigure}[h]{0.325\textwidth}
        \centering
        \includegraphics[width=\textwidth]{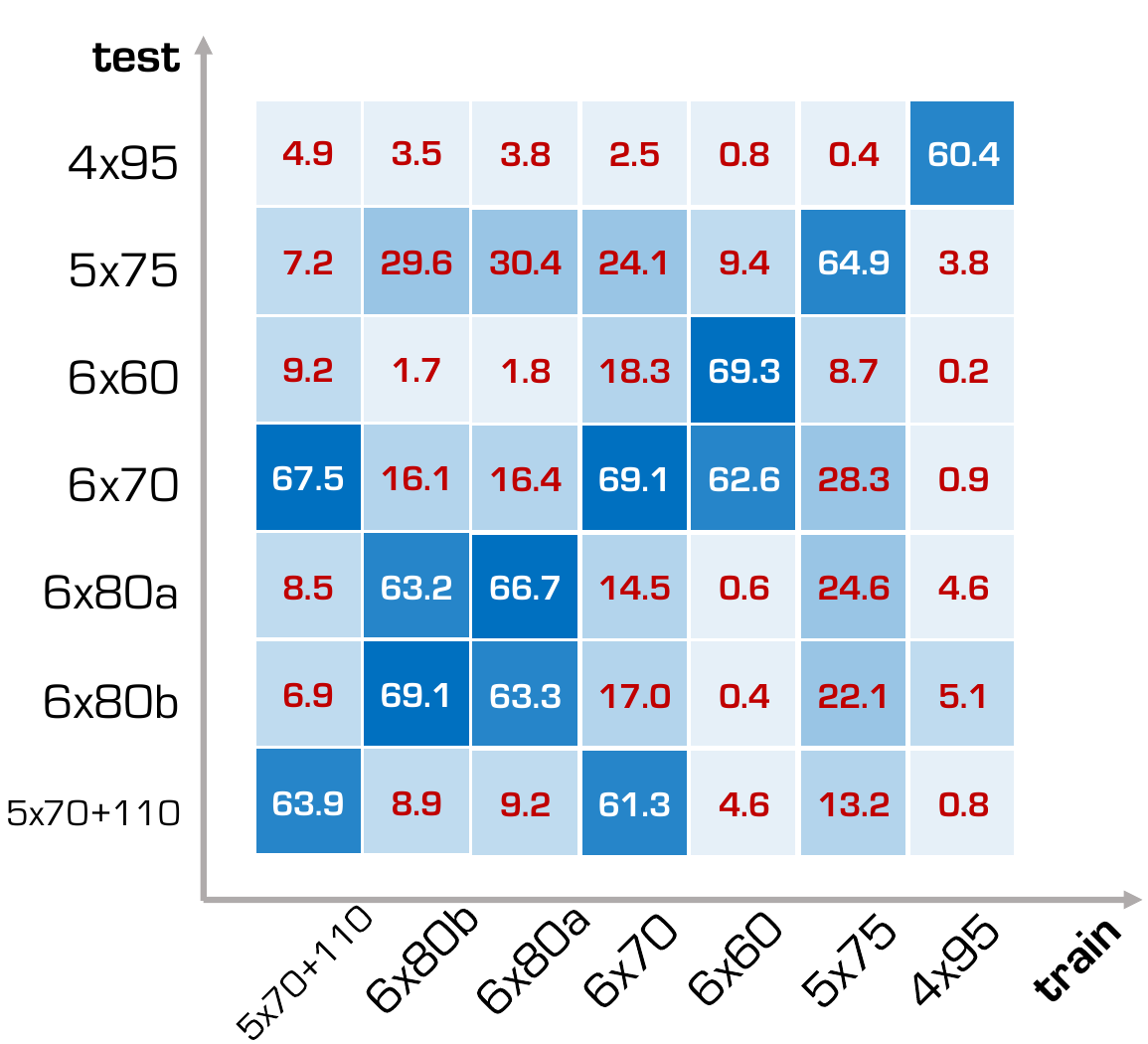}
        \caption{BEVFusion-C~\citep{BEVFusion}}
        \label{}
    \end{subfigure}
    \begin{subfigure}[h]{0.325\textwidth}
        \centering
        \includegraphics[width=\textwidth]{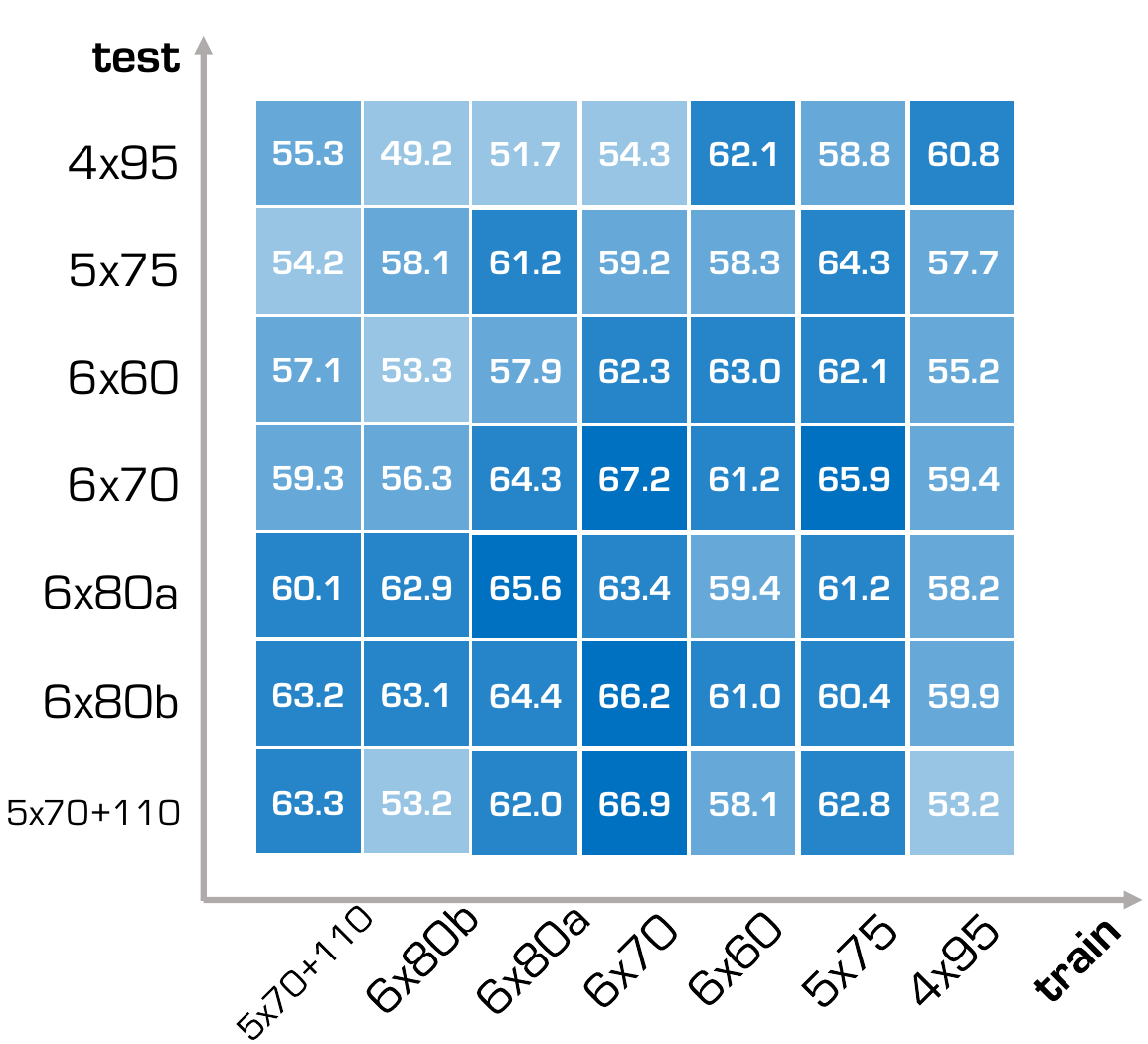}
        \caption{UniDrive (w/o optimization)}
        \label{}
    \end{subfigure}
    \begin{subfigure}[h]{0.325\textwidth}
        \centering
        \includegraphics[width=\textwidth]{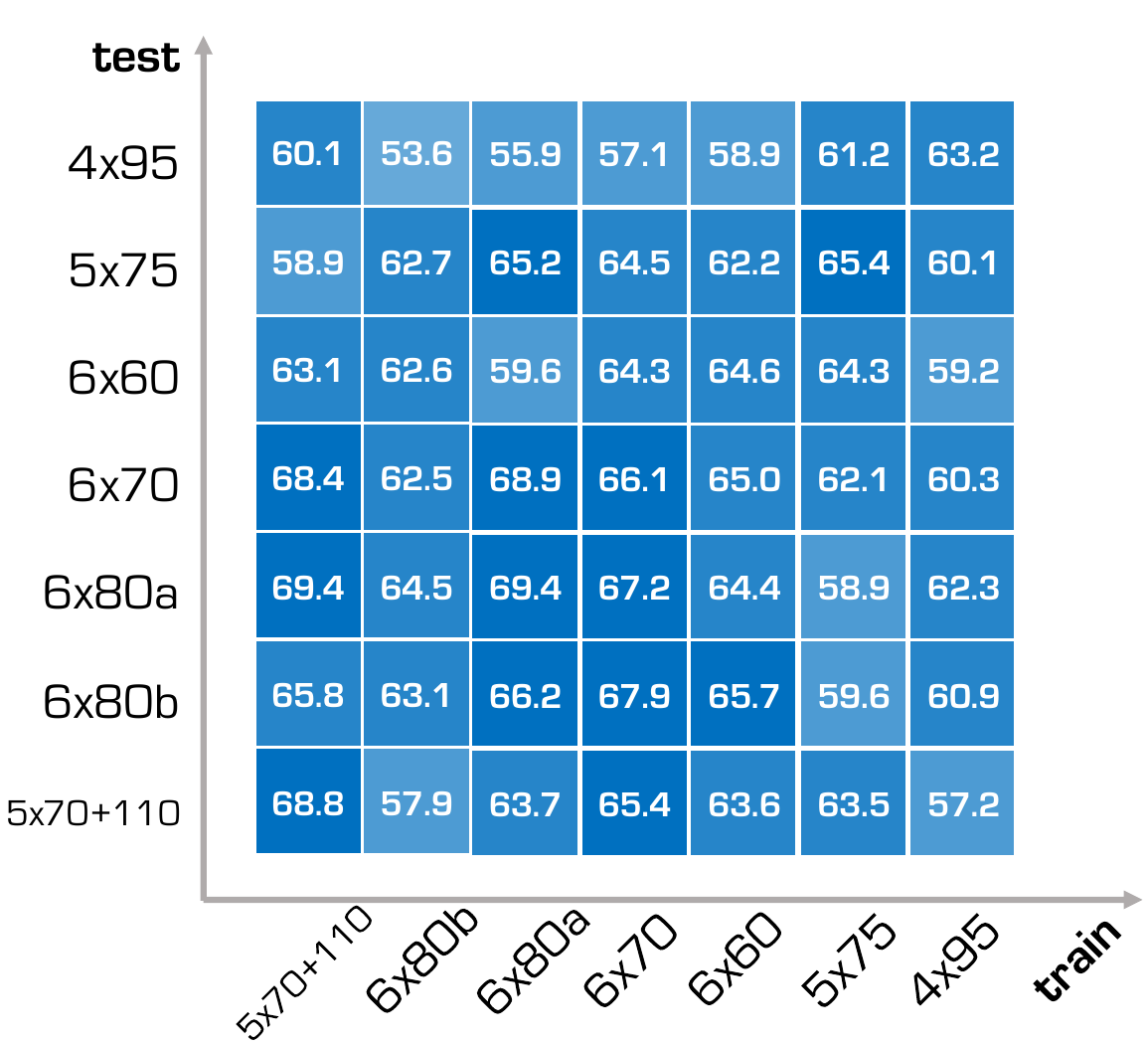}
        \caption{UniDrive (w/ optimization)}
        \label{}
    \end{subfigure}
    \caption{\textbf{Performance evaluations of BEVFusion-C and UniDrive} on 3D object detection across camera configurations. We report the mAP ($\uparrow$) scores in percentage ($\%$). }
    \vspace{-0.2cm}
    \label{fig:map}
\end{figure}

\textbf{Effectiveness of UniDrive.} In Table~\ref{tab:bevfusion} and \ref{tab:unidrive}, we present the 3D object detection results of BEVFusion-C~\citep{BEVFusion} and UniDrive. The models are trained on one configuration and tested on other varying camera configurations. The performance of BEVFusion-C degrades a lot when deployed on cross-camera configuration tasks, nearly unusable on other configurations. As shown in Table~\ref{tab:unidrive}, we train the models using our plug-and-play UniDrive framework. The detection performance significantly improves compared to BEVFusion-C~\citep{BEVFusion}. Our method only experiences little performance degradation on cross-camera configuration tasks. We present more results in Figure~\ref{fig:map}, which comprehensively shows the effectiveness of our framework. 

\textbf{Optimization via UniDrive.} To demonstrate the importance of optimization in UniDrive, we compare the perception performance between optimized virtual camera configurations and intuitive one in Figure~\ref{fig:map}. The intuitive virtual camera configuration places all cameras in the center of the vehicle roof. As shown in Figure~\ref{fig:map} (b), although the intuitive setup (without optimizing) also significantly improved cross-camera configuration perception performance compared to BEVFusion-C (in Figure~\ref{fig:map} (a)), it exhibited a clear preference for certain configurations while performing poorly on others. In contrast, the optimized virtual camera parameters (in Figure~\ref{fig:map} (c)) demonstrated greater adaptability, showing relatively consistent performance across various configurations. This is crucial for the concurrent development of multiple multi-camera perception systems in autonomous driving.

\begin{table}[t]
\centering
\caption{\textbf{Quantitative results of UniDrive for 3D detection across camera configurations}. The detector is trained on the \colorbox{unidrive_blue!8}{\textcolor{unidrive_blue}{blue}} configurations and tested on all configurations directly. We report the mAP ($\uparrow$) and class-level AP ($\uparrow$) scores in percentage ($\%$). }
\label{tab:unidrive}
\renewcommand\arraystretch{1.2}
\setlength\tabcolsep{1pt}
\resizebox{\textwidth}{!}{
\begin{tabular}{c|p{0.90cm}<{\centering}|p{0.90cm}<{\centering}p{0.90cm}<{\centering}p{0.90cm}<{\centering}p{0.90cm}<{\centering}p{0.90cm}<{\centering}p{0.90cm}<{\centering}|c|p{0.90cm}<{\centering}|p{0.90cm}<{\centering}p{0.90cm}<{\centering}p{0.90cm}<{\centering}p{0.90cm}<{\centering}p{0.90cm}<{\centering}p{0.90cm}<{\centering}}
\toprule
\textbf{Configurations} & \textbf{mAP} & \textit{Car} & \textit{Bus} & \textit{Truck} & \textit{Ped.} & \textit{Motor} & \textit{Bic.} & \textbf{Configurations} & \textbf{mAP} & \textit{Car} & \textit{Bus} & \textit{Truck} & \textit{Ped.} & \textit{Motor} & \textit{Bic.} 
\\
\midrule\midrule
\rowcolor{unidrive_blue!8} \textcolor{unidrive_blue}{$5\times70^\circ + 110^\circ$} & \textcolor{unidrive_blue}{68.8} & \textcolor{unidrive_blue}{67.5} & \textcolor{unidrive_blue}{64.8} & \textcolor{unidrive_blue}{71.9} & \textcolor{unidrive_blue}{59.1} & \textcolor{unidrive_blue}{73.6} & \textcolor{unidrive_blue}{75.9} & \textcolor{unidrive_blue}{$6 \times 60^\circ$} & \textcolor{unidrive_blue}{64.6} & \textcolor{unidrive_blue}{63.4} & \textcolor{unidrive_blue}{58.2} & \textcolor{unidrive_blue}{59.7} & \textcolor{unidrive_blue}{59.2} & \textcolor{unidrive_blue}{76.7} & \textcolor{unidrive_blue}{70.0}
\\
\midrule
$4 \times 95^\circ$ & 60.1 & 59.1 & 57.2 & 58.4 & 59.2 & 68.6 & 67.8 & $4 \times 95^\circ$ & 58.9 & 57.7 & 54.1 & 56.9 & 53.1 & 65.2 & 67.4 
\\
$5 \times 75^\circ$& 66.7 & 64.5 & 65.9 & 67.8 & 59.6 & 72.3 & 70.1 & $5 \times 75^\circ$  & 62.2 & 59.5 & 60.6 & 65.8 & 53.6 & 70.3 & 63.1\\
$6 \times 80^\circ a$ & 69.4 & 68.4 & 68.1 & 67.5 & 57.8 & 78.2 & 76.1 & $6 \times 80^\circ a$ & 64.4 & 65.1 & 64.9 & 65.5 & 53.3 & 70.2 & 67.1 
\\
$6 \times 80^\circ b$ & 65.8 & 64.7 & 62.0 & 63.9 & 55.8 & 76.7 & 71.7 & $6 \times 80^\circ b$ & 65.7 & 63.2 & 63.1 & 63.9 & 55.8 & 76.7 & 71.7 
\\
$6 \times 70^\circ$ & 68.4 & 66.8 & 64.3 & 69.8 & 57.6 & 79.1 & 72.8 & $6 \times 70^\circ$ & 65.0 & 62.9 & 60.8 & 65.4 & 55.1 & 73.0 & 72.8 
\\
$6 \times 60^\circ$ & 63.1 & 60.6 & 57.4 & 58.8 & 59.2 & 73.0 & 69.6 & $5\times70^\circ + 110^\circ$ & 63.6 & 61.2 & 61.4 & 64.8 & 55.2 & 71.0 & 68.1 
\\
$8 \times 50^\circ$ & 58.9 & 57.1 & 55.4 & 56.1 & 51.1 & 69.7 & 64.1 & $8 \times 50^\circ$ & 63.8 & 60.2 & 58.3 & 62.6 & 56.8 & 74.2 & 70.5 
\\
\midrule\midrule
\rowcolor{unidrive_blue!8} \textcolor{unidrive_blue}{$6 \times 80^\circ a$} & \textcolor{unidrive_blue}{69.4} & \textcolor{unidrive_blue}{69.0} & \textcolor{unidrive_blue}{67.7} & \textcolor{unidrive_blue}{66.6} & \textcolor{unidrive_blue}{58.6} & \textcolor{unidrive_blue}{78.4} & \textcolor{unidrive_blue}{76.1} & \textcolor{unidrive_blue}{$6 \times 80^\circ b$} & \textcolor{unidrive_blue}{63.1} & \textcolor{unidrive_blue}{63.2} & \textcolor{unidrive_blue}{61.9} & \textcolor{unidrive_blue}{59.8} & \textcolor{unidrive_blue}{53.4} & \textcolor{unidrive_blue}{71.1} & \textcolor{unidrive_blue}{69.4}
\\
\midrule
$4 \times 95^\circ$ & 55.9 & 56.4 & 58.4 & 52.7 & 48.3 & 59.5 & 60.1 & $4 \times 95^\circ$ & 53.6 & 51.2 & 48.0 & 49.7 & 51.0 & 61.5 & 60.1
\\
$5 \times 75^\circ$ & 65.2 & 63.6 & 64.8 & 66.8 & 57.0 & 70.9 & 68.3 & $5 \times 75^\circ$ & 62.7 & 62.1 & 60.6 & 62.2 & 56.3 & 69.9 & 67.0
\\
$5\times70^\circ + 110^\circ$ & 63.7 & 61.3 & 58.9 & 60.5 & 59.9 & 72.8 & 68.8 & $6 \times 80^\circ a$ & 64.5 & 62.6 & 60.3 & 62.4 & 58.6 & 71.4 & 71.6
\\
$6 \times 80^\circ b$ & 66.2 & 65.2 & 61.1 & 65.1 & 56.1 & 76.3 & 73.2 &  $6 \times 60^\circ$ & 62.6 & 60.3 & 59.1 & 62.6 & 53.4 & 70.2 & 70.2
\\
$6 \times 70^\circ$ & 68.9 & 67.9 & 63.5 & 70.7 & 58.3 & 79.5 & 73.8 & $6 \times 70^\circ$ & 62.5 & 59.4 & 55.6 & 62.9 & 52.6 & 73.4 & 70.8
\\
$6 \times 60^\circ$ & 59.6 & 57.2 & 54.0 & 55.7 & 57.0 & 67.5 & 66.1 & $5\times70^\circ + 110^\circ$ & 57.9 & 56.1 & 52.5 & 53.7 & 56.9 & 64.3 & 63.7
\\
$8 \times 50^\circ$ & 61.2 & 60.3 & 58.1 & 59.9 & 54.5 & 68.2 & 65.9 & $8 \times 50^\circ$ & 58.4 & 60.3 & 57.0 & 54.3 & 53.9 & 64.4 & 60.2
\\
\bottomrule
\end{tabular}
}
\vspace{-1mm}
\end{table}

\subsection{Ablation Study}

In this section, we analyze the interplay between our proposed virtual projection strategy and perception performance to address these questions: \textit{1) What's the impact of camera extrinsic and intrinsic for cross-configuration perception? 2) How UniDrive works towards these parameters separately?}

\begin{figure}[t]
    \begin{subfigure}[h]{0.344\textwidth}
        \centering
        \includegraphics[width=\textwidth]{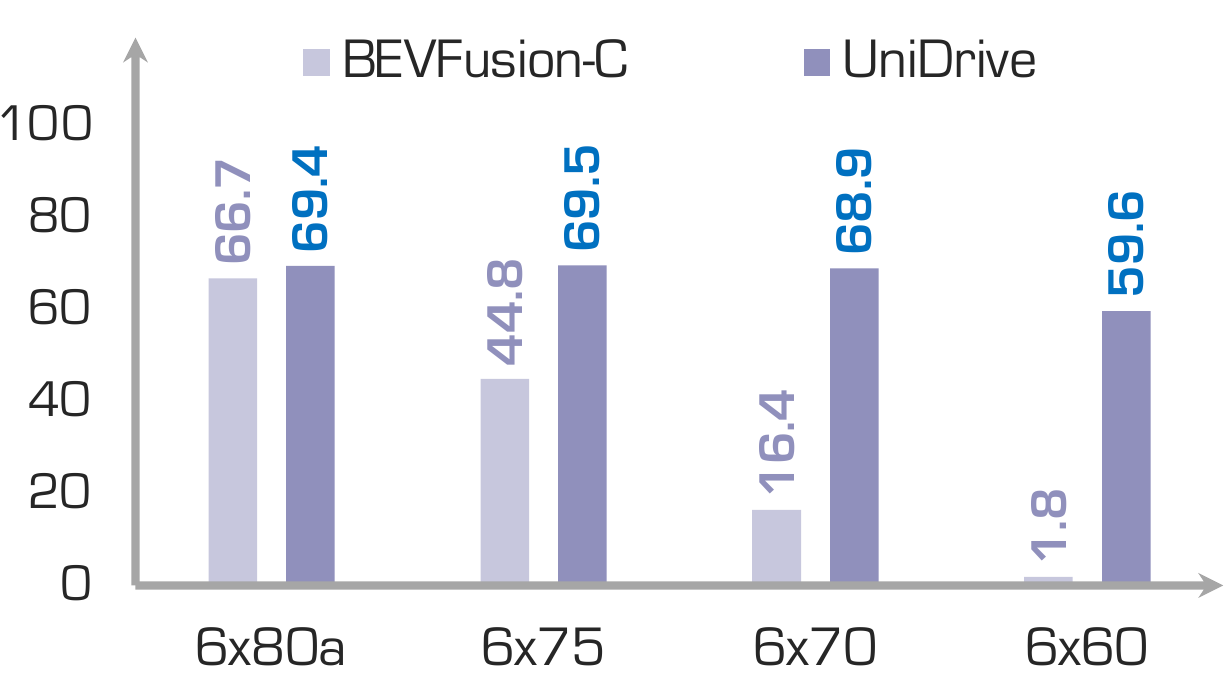}
        \caption{Camera intrinsic}
        \label{}
    \end{subfigure}
    \begin{subfigure}[h]{0.344\textwidth}
        \centering
        \includegraphics[width=\textwidth]{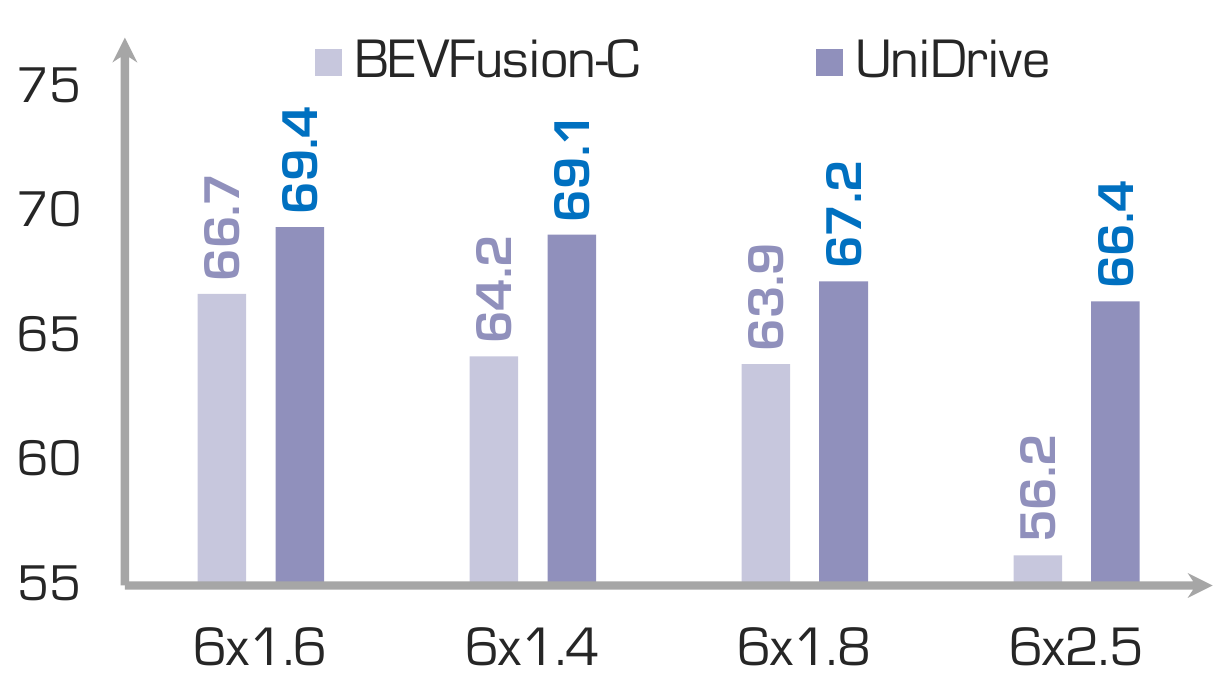}
        \caption{Camera height}
        \label{}
    \end{subfigure}
    \begin{subfigure}[h]{0.27\textwidth}
        \centering
        \includegraphics[width=\textwidth]{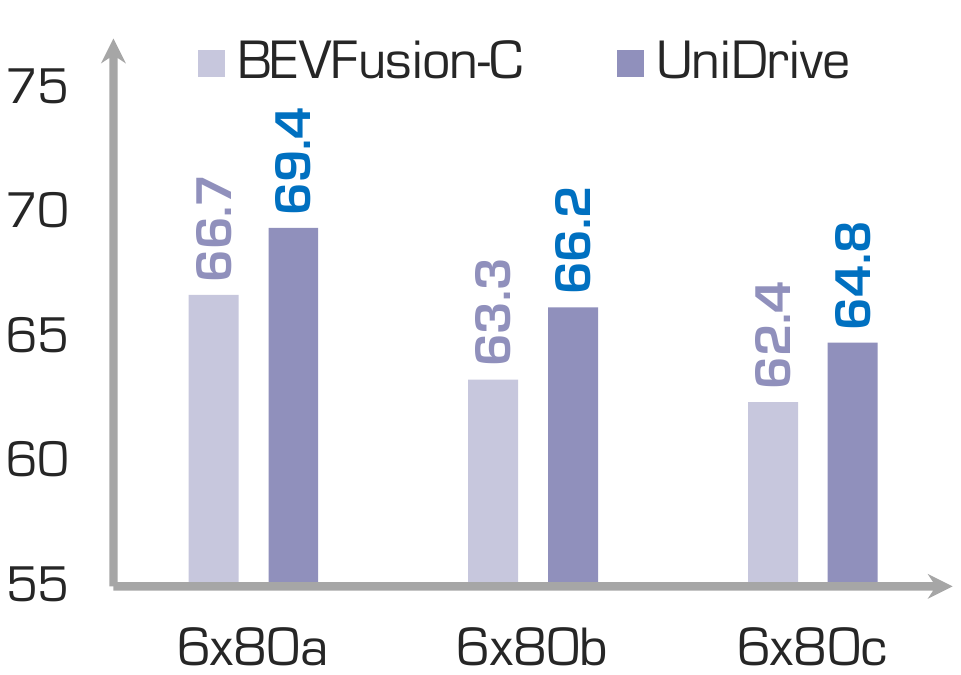}
        \caption{Camera placement}
        \label{}
    \end{subfigure}
    \caption{\textbf{Ablation Study of BEVFusion-C and UniDrive} on 3D object detection across camera configurations. We report the mAP ($\uparrow$) scores in percentage ($\%$). }
    \vspace{-0.2cm}
    \label{fig:ablation}
\end{figure}

\textbf{Camera Intrinsics.} Changes in camera intrinsics pose the greatest challenge for cross-camera parameter perception. In Figure~\ref{fig:map} (a), BEVFusion-C almost entirely fails when tasked on distinct camera intrinsics with the detection accuracy mostly under \textit{20\%}. For instance, BEVFusion only gets \textit{1.8\%} when deploying models trained on $6 \times 80^\circ a$ to $6 \times 60^\circ$. In contrast, in Figure~\ref{fig:ablation}, our UniDrive framework demonstrates substantial robustness, with performance dropping by at most \textit{9.8\%} under the largest intrinsic differences, which highlights the effectiveness of our approach.

\textbf{Camera Height.} The variation in the vertical position of cameras can significantly impact perception performance, as cameras at varying heights capture images with distinct geometric features. 
We perform experiments specifically for varying camera heights at 1.6 meters, 1.4 meters, 1.8 meters, and 2.5 meters. We train the model on 1.6 meters and test on other configuraitons. As shown in Fig.~\ref{fig:ablation}b. BEVFusion-C experiences a substantial performance drop for more than \textit{10\%}, when faced with varying camera heights. In contrast, UniDrive significantly improves performance across different camera heights, demonstrating enhanced robustness with only \textit{3.0\%} performance decreasing.

\textbf{Camera Placement.} Changing the camera's horizontal position and orientation on presents a relatively smaller challenge for cross-camera parameter perception. As shown in Figure~\ref{fig:map} (a), BEVFusion-C experiences a performance drop of \textit{5.9\%} when deploying the model trained on the $6 \times 80^\circ b$ configuration to the $6 \times 80^\circ a$ configuration. Nonetheless, our UniDrive framework further enhances cross-camera parameter perception performance. In Figure~\ref{fig:map} (c), we train the model on the $6 \times 80^\circ a$ configuration and test on other configurations, UniDrive only experiences a \textit{4.6\%} when deploying the model trained on the $6 \times 80^\circ b$ configuration to the $6 \times 80^\circ a$ configuration.

\subsection{Analysis}

In this section, we further investigate some useful insight points found in the benchmark experiments: \textit{1) What's the impact of inconsistency in multi-camera intrinsics for perception? 2) How UniDrive works towards this inconsistency?}

\textbf{Degradation with Inconsistent Intrinsics.} In our experiments, we observed that for multi-camera systems, models perform better when camera intrinsics are consistent compared to when they vary. However, due to design aesthetics and other constraints, many autonomous driving companies use multiple cameras with different intrinsic parameters to achieve 360-degree perception. For instance, the nuScenes~\citep{nuScenes} uses five $70^\circ$ cameras and one $110^\circ$ camera. As shown in Fig~\ref{tab:bevfusion}, BEVFusion-C performs a lot better in $6 \times 80^\circ a$ and $6 \times 60^\circ$ compared to configuration $5\times70^\circ+110^\circ$. Thus, inconsistency in camera intrinsics can potentially hinder perception improvement. 

\textbf{Improvement via UniDrive.} Our framework significantly enhances the perception performance of multi-camera systems with varying intrinsics by leveraging a virtual camera system with consistent intrinsics. For training and testing on the same configurations, as demonstrated in Figure~\ref{tab:unidrive}, UniDrvie achieves \textit{68.8\%} accuracy in $5\times70^\circ + 110^\circ$ configuration, which surpasses \textit{4.9\%} than BEVFusion-C (\textit{63.9\%}). For testing across camera configurations, UniDrive experiences little accuracy reduction only in rare situations. This demonstrates that UniDrive has substantial potential to push advancements in driving perception technology.

\section{Conclusion}

\vspace{-2mm}

In this paper, we introduce the UniDrive framework, a robust solution for enhancing the generalization of vision-centric autonomous driving models across varying camera configurations. By leveraging a unified set of virtual cameras and a ground-aware projection method, our approach effectively mitigates the challenges posed by camera intrinsics and extrinsics. The proposed virtual configuration optimization ensures minimal projection error, enabling adaptable and reliable performance across diverse sensor setups. Extensive experiments in CARLA validate the effectiveness of UniDrive, demonstrating strong generalization capabilities with minimal performance loss. Our framework not only serves as a plug-and-play module for existing 3D perception models but also paves the way for more versatile and scalable autonomous driving solutions.

\textbf{Limitation.} The camera configurations analyzed in this paper can not cover all real-world setups, more comprehensive experiments may be required. In addition, our research are fully conducted on simulation data, as real-world experiments are time-consuming and need extensive resource.

\clearpage



\bibliography{main}

\begin{thebibliography}{57}
\providecommand{\natexlab}[1]{#1}
\providecommand{\url}[1]{\texttt{#1}}
\expandafter\ifx\csname urlstyle\endcsname\relax
  \providecommand{\doi}[1]{doi: #1}\else
  \providecommand{\doi}{doi: \begingroup \urlstyle{rm}\Url}\fi

\bibitem[Caesar et~al.(2020)Caesar, Bankiti, Lang, Vora, Liong, Xu, Krishnan, Pan, Baldan, and Beijbom]{nuScenes}
Holger Caesar, Varun Bankiti, Alex~H Lang, Sourabh Vora, Venice~Erin Liong, Qiang Xu, Anush Krishnan, Yu~Pan, Giancarlo Baldan, and Oscar Beijbom.
\newblock nuscenes: A multimodal dataset for autonomous driving.
\newblock In \emph{CVPR}, pp.\  11621--11631, 2020.

\bibitem[Cai et~al.(2023)Cai, Jiang, Xu, Zhao, Ma, Liu, and Li]{cai2023analyzing}
Xinyu Cai, Wentao Jiang, Runsheng Xu, Wenquan Zhao, Jiaqi Ma, Si~Liu, and Yikang Li.
\newblock Analyzing infrastructure lidar placement with realistic lidar simulation library.
\newblock In \emph{ICRA}, pp.\  5581--5587, 2023.

\bibitem[Chen et~al.(2022{\natexlab{a}})Chen, Wang, Wang, Tian, Xiong, and Li]{chen2022epro}
Hansheng Chen, Pichao Wang, Fan Wang, Wei Tian, Lu~Xiong, and Hao Li.
\newblock Epro-pnp: Generalized end-to-end probabilistic perspective-n-points for monocular object pose estimation.
\newblock In \emph{CVPR}, pp.\  2781--2790, 2022{\natexlab{a}}.

\bibitem[Chen et~al.(2018)Chen, Li, Sakaridis, Dai, and Van~Gool]{chen2018domain}
Yuhua Chen, Wen Li, Christos Sakaridis, Dengxin Dai, and Luc Van~Gool.
\newblock Domain adaptive faster r-cnn for object detection in the wild.
\newblock In \emph{CVPR}, pp.\  3339--3348, 2018.

\bibitem[Chen et~al.(2022{\natexlab{b}})Chen, Li, Zhang, Fang, Jiang, and Zhao]{chen2022graph}
Zehui Chen, Zhenyu Li, Shiquan Zhang, Liangji Fang, Qinhong Jiang, and Feng Zhao.
\newblock Graph-detr3d: rethinking overlapping regions for multi-view 3d object detection.
\newblock In \emph{ACM MM}, pp.\  5999--6008, 2022{\natexlab{b}}.

\bibitem[Dosovitskiy et~al.(2017)Dosovitskiy, Ros, Codevilla, Lopez, and Koltun]{Dosovitskiy17}
Alexey Dosovitskiy, German Ros, Felipe Codevilla, Antonio Lopez, and Vladlen Koltun.
\newblock {CARLA}: {An} open urban driving simulator.
\newblock In \emph{CoRL}, pp.\  1--16, 2017.

\bibitem[Dou et~al.(2019)Dou, Coelho~de Castro, Kamnitsas, and Glocker]{dou2019domain}
Qi~Dou, Daniel Coelho~de Castro, Konstantinos Kamnitsas, and Ben Glocker.
\newblock Domain generalization via model-agnostic learning of semantic features.
\newblock \emph{NeurIPS}, 32, 2019.

\bibitem[Facil et~al.(2019)Facil, Ummenhofer, Zhou, Montesano, Brox, and Civera]{facil2019cam}
Jose~M Facil, Benjamin Ummenhofer, Huizhong Zhou, Luis Montesano, Thomas Brox, and Javier Civera.
\newblock Cam-convs: Camera-aware multi-scale convolutions for single-view depth.
\newblock In \emph{CVPR}, pp.\  11826--11835, 2019.

\bibitem[Hansen(2016)]{hansen2016cma}
Nikolaus Hansen.
\newblock The cma evolution strategy: A tutorial.
\newblock \emph{arXiv preprint arXiv:1604.00772}, 2016.

\bibitem[Hao et~al.(2024)Hao, Wei, Yang, Zhao, Zhang, Zhou, Wang, Li, Kong, and Zhang]{hao2024your}
Xiaoshuai Hao, Mengchuan Wei, Yifan Yang, Haimei Zhao, Hui Zhang, Yi~Zhou, Qiang Wang, Weiming Li, Lingdong Kong, and Jing Zhang.
\newblock Is your hd map constructor reliable under sensor corruptions?
\newblock \emph{arXiv preprint arXiv:2406.12214}, 2024.

\bibitem[He \& Zhang(2020)He and Zhang]{he2020domain}
Zhenwei He and Lei Zhang.
\newblock Domain adaptive object detection via asymmetric tri-way faster-rcnn.
\newblock In \emph{ECCV}, pp.\  309--324, 2020.

\bibitem[Hu et~al.(2022)Hu, Liu, Chitlangia, Agnihotri, and Zhao]{hu2022investigating}
Hanjiang Hu, Zuxin Liu, Sharad Chitlangia, Akhil Agnihotri, and Ding Zhao.
\newblock Investigating the impact of multi-lidar placement on object detection for autonomous driving.
\newblock In \emph{CVPR}, pp.\  2550--2559, 2022.

\bibitem[Hu et~al.(2023)Hu, Yang, Chen, Li, Sima, Zhu, Chai, Du, Lin, Wang, et~al.]{hu2023uniAD}
Yihan Hu, Jiazhi Yang, Li~Chen, Keyu Li, Chonghao Sima, Xizhou Zhu, Siqi Chai, Senyao Du, Tianwei Lin, Wenhai Wang, et~al.
\newblock Planning-oriented autonomous driving.
\newblock In \emph{CVPR}, pp.\  17853--17862, 2023.

\bibitem[Huang \& Huang(2022)Huang and Huang]{huang2022bevdet4d}
Junjie Huang and Guan Huang.
\newblock Bevdet4d: Exploit temporal cues in multi-camera 3d object detection.
\newblock \emph{arXiv preprint arXiv:2203.17054}, 2022.

\bibitem[Huang et~al.(2021)Huang, Huang, Zhu, Ye, and Du]{huang2021bevdet}
Junjie Huang, Guan Huang, Zheng Zhu, Yun Ye, and Dalong Du.
\newblock Bevdet: High-performance multi-camera 3d object detection in bird-eye-view.
\newblock \emph{arXiv preprint arXiv:2112.11790}, 2021.

\bibitem[Huang et~al.(2023)Huang, Zheng, Zhang, Zhou, and Lu]{huang2023tri}
Yuanhui Huang, Wenzhao Zheng, Yunpeng Zhang, Jie Zhou, and Jiwen Lu.
\newblock Tri-perspective view for vision-based 3d semantic occupancy prediction.
\newblock In \emph{CVPR}, pp.\  9223--9232, 2023.

\bibitem[Huang et~al.(2024{\natexlab{a}})Huang, Zheng, Zhang, Zhou, and Lu]{huang2024selfocc}
Yuanhui Huang, Wenzhao Zheng, Borui Zhang, Jie Zhou, and Jiwen Lu.
\newblock Selfocc: Self-supervised vision-based 3d occupancy prediction.
\newblock In \emph{CVPR}, pp.\  19946--19956, 2024{\natexlab{a}}.

\bibitem[Huang et~al.(2024{\natexlab{b}})Huang, Zheng, Zhang, Zhou, and Lu]{huang2024gaussianformer}
Yuanhui Huang, Wenzhao Zheng, Yunpeng Zhang, Jie Zhou, and Jiwen Lu.
\newblock Gaussianformer: Scene as gaussians for vision-based 3d semantic occupancy prediction.
\newblock \emph{arXiv preprint arXiv:2405.17429}, 2024{\natexlab{b}}.

\bibitem[Jiang et~al.(2023{\natexlab{a}})Jiang, Chen, Xu, Liao, Chen, Zhou, Zhang, Liu, Huang, and Wang]{jiang2023vad}
Bo~Jiang, Shaoyu Chen, Qing Xu, Bencheng Liao, Jiajie Chen, Helong Zhou, Qian Zhang, Wenyu Liu, Chang Huang, and Xinggang Wang.
\newblock Vad: Vectorized scene representation for efficient autonomous driving.
\newblock \emph{arXiv preprint arXiv:2303.12077}, 2023{\natexlab{a}}.

\bibitem[Jiang et~al.(2023{\natexlab{b}})Jiang, Xiang, Cai, Xu, Ma, Li, Lee, and Liu]{jiang2023optimizing}
Wentao Jiang, Hao Xiang, Xinyu Cai, Runsheng Xu, Jiaqi Ma, Yikang Li, Gim~Hee Lee, and Si~Liu.
\newblock Optimizing the placement of roadside lidars for autonomous driving.
\newblock In \emph{ICCV}, pp.\  18381--18390, 2023{\natexlab{b}}.

\bibitem[Jin et~al.(2022)Jin, Gao, Hui, Zhao, Wei, Ma, and Gan]{jin2022roadside}
Shaojie Jin, Ying Gao, Fei Hui, Xiangmo Zhao, Cheng Wei, Tao Ma, and Weihao Gan.
\newblock A novel information theory-based metric for evaluating roadside lidar placement.
\newblock \emph{IEEE Sensors Journal}, 22\penalty0 (21):\penalty0 21009--21023, 2022.

\bibitem[Joshi \& Boyd(2008)Joshi and Boyd]{joshi2008sensor}
Siddharth Joshi and Stephen Boyd.
\newblock Sensor selection via convex optimization.
\newblock \emph{IEEE Transactions on Signal Processing}, 57\penalty0 (2):\penalty0 451--462, 2008.

\bibitem[Kim et~al.(2023)Kim, Jo, Yun, Yun, and Park]{kim2023placement}
Tae-Hyeong Kim, Gi-Hwan Jo, Hyeong-Seok Yun, Kyung-Su Yun, and Tae-Hyoung Park.
\newblock Placement method of multiple lidars for roadside infrastructure in urban environments.
\newblock \emph{Sensors}, 23\penalty0 (21):\penalty0 8808, 2023.

\bibitem[Li et~al.(2018)Li, Pan, Wang, and Kot]{li2018domain}
Haoliang Li, Sinno~Jialin Pan, Shiqi Wang, and Alex~C Kot.
\newblock Domain generalization with adversarial feature learning.
\newblock In \emph{CVPR}, pp.\  5400--5409, 2018.

\bibitem[Li et~al.(2024{\natexlab{a}})Li, Hu, Liu, Xu, Huang, and Zhao]{li2024influence}
Ye~Li, Hanjiang Hu, Zuxin Liu, Xiaohao Xu, Xiaonan Huang, and Ding Zhao.
\newblock Influence of camera-lidar configuration on 3d object detection for autonomous driving.
\newblock In \emph{ICRA}, pp.\  9018--9025, 2024{\natexlab{a}}.

\bibitem[Li et~al.(2024{\natexlab{b}})Li, Kong, Hu, Xu, and Huang]{li2024your}
Ye~Li, Lingdong Kong, Hanjiang Hu, Xiaohao Xu, and Xiaonan Huang.
\newblock Is your lidar placement optimized for 3d scene understanding?
\newblock In \emph{NeurIPS}, 2024{\natexlab{b}}.

\bibitem[Li et~al.(2022)Li, Wang, Li, Xie, Sima, Lu, Qiao, and Dai]{li2022bevformer}
Zhiqi Li, Wenhai Wang, Hongyang Li, Enze Xie, Chonghao Sima, Tong Lu, Yu~Qiao, and Jifeng Dai.
\newblock Bevformer: Learning bird’s-eye-view representation from multi-camera images via spatiotemporal transformers.
\newblock In \emph{ECCV}, pp.\  1--18, 2022.

\bibitem[Liu et~al.(2024)Liu, Huang, Zhang, Yao, Zhang, Wan, Ye, and Zhou]{liu2024ray}
Feng Liu, Tengteng Huang, Qianjing Zhang, Haotian Yao, Chi Zhang, Fang Wan, Qixiang Ye, and Yanzhao Zhou.
\newblock Ray denoising: Depth-aware hard negative sampling for multi-view 3d object detection.
\newblock In \emph{ECCV}, 2024.

\bibitem[Liu et~al.(2022)Liu, Wang, Zhang, and Sun]{liu2022petr}
Yingfei Liu, Tiancai Wang, Xiangyu Zhang, and Jian Sun.
\newblock Petr: Position embedding transformation for multi-view 3d object detection.
\newblock In \emph{ECCV}, pp.\  531--548, 2022.

\bibitem[Liu et~al.(2023{\natexlab{a}})Liu, Yan, Jia, Li, Gao, Wang, and Zhang]{liu2023petrv2}
Yingfei Liu, Junjie Yan, Fan Jia, Shuailin Li, Aqi Gao, Tiancai Wang, and Xiangyu Zhang.
\newblock Petrv2: A unified framework for 3d perception from multi-camera images.
\newblock In \emph{ICCV}, pp.\  3262--3272, 2023{\natexlab{a}}.

\bibitem[Liu et~al.(2023{\natexlab{b}})Liu, Tang, Amini, Yang, Mao, Rus, and Han]{BEVFusion}
Zhijian Liu, Haotian Tang, Alexander Amini, Xinyu Yang, Huizi Mao, Daniela~L. Rus, and Song Han.
\newblock Bevfusion: Multi-task multi-sensor fusion with unified bird's-eye view representation.
\newblock In \emph{ICRA}, pp.\  2774--2781, 2023{\natexlab{b}}.

\bibitem[Liu et~al.(2019)Liu, Arief, and Zhao]{liu2019should}
Zuxin Liu, Mansur Arief, and Ding Zhao.
\newblock Where should we place lidars on the autonomous vehicle?-an optimal design approach.
\newblock In \emph{ICRA}, pp.\  2793--2799, 2019.

\bibitem[Lu et~al.(2022)Lu, Zhou, Zhu, Xu, and Zhang]{lu2022learning}
Jiachen Lu, Zheyuan Zhou, Xiatian Zhu, Hang Xu, and Li~Zhang.
\newblock Learning ego 3d representation as ray tracing.
\newblock In \emph{ECCV}, pp.\  129--144, 2022.

\bibitem[Muandet et~al.(2013)Muandet, Balduzzi, and Sch{\"o}lkopf]{muandet2013domain}
Krikamol Muandet, David Balduzzi, and Bernhard Sch{\"o}lkopf.
\newblock Domain generalization via invariant feature representation.
\newblock In \emph{ICML}, pp.\  10--18, 2013.

\bibitem[Peng et~al.(2023)Peng, Chen, Qiao, Kong, Liu, Wang, Zhu, and Ma]{peng2023learning}
Xidong Peng, Runnan Chen, Feng Qiao, Lingdong Kong, Youquan Liu, T~Wang, X~Zhu, and Y~Ma.
\newblock Learning to adapt sam for segmenting cross-domain point clouds.
\newblock \emph{arXiv preprint arXiv:2310.08820}, 2023.

\bibitem[Philion \& Fidler(2020)Philion and Fidler]{philion2020lift}
Jonah Philion and Sanja Fidler.
\newblock Lift, splat, shoot: Encoding images from arbitrary camera rigs by implicitly unprojecting to 3d.
\newblock In \emph{ECCV}, 2020.

\bibitem[Reading et~al.(2021)Reading, Harakeh, Chae, and Waslander]{reading2021categorical}
Cody Reading, Ali Harakeh, Julia Chae, and Steven~L Waslander.
\newblock Categorical depth distribution network for monocular 3d object detection.
\newblock In \emph{CVPR}, pp.\  8555--8564, 2021.

\bibitem[Sun et~al.(2020)Sun, Kretzschmar, Dotiwalla, Chouard, Patnaik, Tsui, Guo, Zhou, Chai, Caine, Vasudevan, Han, Ngiam, Zhao, Timofeev, Ettinger, Krivokon, Gao, Joshi, Zhang, Shlens, Chen, and Anguelov]{sun2020waymoOpen}
Pei Sun, Henrik Kretzschmar, Xerxes Dotiwalla, Aurelien Chouard, Vijaysai Patnaik, Paul Tsui, James Guo, Yin Zhou, Yuning Chai, Benjamin Caine, Vijay Vasudevan, Wei Han, Jiquan Ngiam, Hang Zhao, Aleksei Timofeev, Scott Ettinger, Maxim Krivokon, Amy Gao, Aditya Joshi, Yu~Zhang, Jonathon Shlens, Zhifeng Chen, and Dragomir Anguelov.
\newblock Scalability in perception for autonomous driving: Waymo open dataset.
\newblock In \emph{CVPR}, pp.\  2446--2454, 2020.

\bibitem[Tian et~al.(2020)Tian, Shen, Chen, and He]{tian2020fcos}
Zhi Tian, Chunhua Shen, Hao Chen, and Tong He.
\newblock Fcos: A simple and strong anchor-free object detector.
\newblock \emph{TPAMI}, 44\penalty0 (4):\penalty0 1922--1933, 2020.

\bibitem[Wang et~al.(2023{\natexlab{a}})Wang, Liu, Wang, Li, and Zhang]{wang2023exploring}
Shihao Wang, Yingfei Liu, Tiancai Wang, Ying Li, and Xiangyu Zhang.
\newblock Exploring object-centric temporal modeling for efficient multi-view 3d object detection.
\newblock In \emph{ICCV}, pp.\  3621--3631, 2023{\natexlab{a}}.

\bibitem[Wang et~al.(2023{\natexlab{b}})Wang, Zhao, Xu, Chen, Yu, Chang, Yang, and Zhao]{wang2023towards}
Shuo Wang, Xinhai Zhao, Hai-Ming Xu, Zehui Chen, Dameng Yu, Jiahao Chang, Zhen Yang, and Feng Zhao.
\newblock Towards domain generalization for multi-view 3d object detection in bird-eye-view.
\newblock In \emph{CVPR}, pp.\  13333--13342, 2023{\natexlab{b}}.

\bibitem[Wang et~al.(2021)Wang, Zhu, Pang, and Lin]{wang2021fcos3d}
Tai Wang, Xinge Zhu, Jiangmiao Pang, and Dahua Lin.
\newblock Fcos3d: Fully convolutional one-stage monocular 3d object detection.
\newblock In \emph{ICCV}, pp.\  913--922, 2021.

\bibitem[Wang et~al.(2022{\natexlab{a}})Wang, Xinge, Pang, and Lin]{wang2022probabilistic}
Tai Wang, ZHU Xinge, Jiangmiao Pang, and Dahua Lin.
\newblock Probabilistic and geometric depth: Detecting objects in perspective.
\newblock In \emph{CoRL}, pp.\  1475--1485, 2022{\natexlab{a}}.

\bibitem[Wang et~al.(2022{\natexlab{b}})Wang, Zhang, Yang, and Sun]{wang2022anchor}
Yingming Wang, Xiangyu Zhang, Tong Yang, and Jian Sun.
\newblock Anchor detr: Query design for transformer-based detector.
\newblock In \emph{Proceedings of the AAAI conference on artificial intelligence}, 2022{\natexlab{b}}.

\bibitem[Wang et~al.(2022{\natexlab{c}})Wang, Guizilini, Zhang, Wang, Zhao, and Solomon]{wang2022detr3d}
Yue Wang, Vitor~Campagnolo Guizilini, Tianyuan Zhang, Yilun Wang, Hang Zhao, and Justin Solomon.
\newblock Detr3d: 3d object detection from multi-view images via 3d-to-2d queries.
\newblock In \emph{CoRL}, pp.\  180--191, 2022{\natexlab{c}}.

\bibitem[Wei et~al.(2023)Wei, Zhao, Zheng, Zhu, Zhou, and Lu]{wei2023surroundocc}
Yi~Wei, Linqing Zhao, Wenzhao Zheng, Zheng Zhu, Jie Zhou, and Jiwen Lu.
\newblock Surroundocc: Multi-camera 3d occupancy prediction for autonomous driving.
\newblock In \emph{ICCV}, pp.\  21729--21740, 2023.

\bibitem[Xie et~al.(2022)Xie, Yu, Zhou, Philion, Anandkumar, Fidler, Luo, and Alvarez]{xie2204m2bev}
E~Xie, Z~Yu, D~Zhou, J~Philion, A~Anandkumar, S~Fidler, P~Luo, and JM~Alvarez.
\newblock M2bev: Multi-camera joint 3d detection and segmentation with unified birds-eye view representation.
\newblock \emph{arXiv preprint arXiv:2204.05088}, 2022.

\bibitem[Xu et~al.(2020)Xu, Zhao, Jin, and Wei]{xu2020exploring}
Chang-Dong Xu, Xing-Ran Zhao, Xin Jin, and Xiu-Shen Wei.
\newblock Exploring categorical regularization for domain adaptive object detection.
\newblock In \emph{CVPR}, pp.\  11724--11733, 2020.

\bibitem[Xu et~al.(2022)Xu, Du, Zhang, Zhang, Hong, Huang, and Han]{xu2022optimization}
Xiaohao Xu, Zihao Du, Huaxin Zhang, Ruichao Zhang, Zihan Hong, Qin Huang, and Bin Han.
\newblock Optimization of forcemyography sensor placement for arm movement recognition.
\newblock In \emph{IROS}, pp.\  9845--9850, 2022.

\bibitem[Zeng et~al.(2024)Zeng, Zheng, Lu, and Yan]{zeng2024hardness}
Shuai Zeng, Wenzhao Zheng, Jiwen Lu, and Haibin Yan.
\newblock Hardness-aware scene synthesis for semi-supervised 3d object detection.
\newblock \emph{TMM}, 2024.

\bibitem[Zhao et~al.(2020)Zhao, Li, Xu, and Lin]{zhao2020collaborative}
Ganlong Zhao, Guanbin Li, Ruijia Xu, and Liang Lin.
\newblock Collaborative training between region proposal localization and classification for domain adaptive object detection.
\newblock In \emph{ECCV}, pp.\  86--102, 2020.

\bibitem[Zhao et~al.(2024)Zhao, Xu, Wang, Zhang, Zhang, Zheng, Du, Zhou, and Lu]{zhao2024lowrankocc}
Linqing Zhao, Xiuwei Xu, Ziwei Wang, Yunpeng Zhang, Borui Zhang, Wenzhao Zheng, Dalong Du, Jie Zhou, and Jiwen Lu.
\newblock Lowrankocc: Tensor decomposition and low-rank recovery for vision-based 3d semantic occupancy prediction.
\newblock In \emph{CVPR}, pp.\  9806--9815, 2024.

\bibitem[Zheng et~al.(2024{\natexlab{a}})Zheng, Chen, Huang, Zhang, Duan, and Lu]{zheng2023occworld}
Wenzhao Zheng, Weiliang Chen, Yuanhui Huang, Borui Zhang, Yueqi Duan, and Jiwen Lu.
\newblock Occworld: Learning a 3d occupancy world model for autonomous driving.
\newblock In \emph{ECCV}, 2024{\natexlab{a}}.

\bibitem[Zheng et~al.(2024{\natexlab{b}})Zheng, Song, Guo, and Chen]{zheng2024genad}
Wenzhao Zheng, Ruiqi Song, Xianda Guo, and Long Chen.
\newblock Genad: Generative end-to-end autonomous driving.
\newblock In \emph{ECCV}, 2024{\natexlab{b}}.

\bibitem[Zhou \& Kr{\"a}henb{\"u}hl(2022)Zhou and Kr{\"a}henb{\"u}hl]{zhou2022cross}
Brady Zhou and Philipp Kr{\"a}henb{\"u}hl.
\newblock Cross-view transformers for real-time map-view semantic segmentation.
\newblock In \emph{CVPR}, pp.\  13760--13769, 2022.

\bibitem[Zhu et~al.(2021)Zhu, Su, Lu, Li, Wang, and Dai]{zhu2021deformable}
Xizhou Zhu, Weijie Su, Lewei Lu, Bin Li, Xiaogang Wang, and Jifeng Dai.
\newblock Deformable detr: Deformable transformers for end-to-end object detection.
\newblock In \emph{ICLR}, 2021.

\bibitem[Zong et~al.(2023)Zong, Jiang, Song, Xue, Su, Li, and Liu]{zong2023temporal}
Zhuofan Zong, Dongzhi Jiang, Guanglu Song, Zeyue Xue, Jingyong Su, Hongsheng Li, and Yu~Liu.
\newblock Temporal enhanced training of multi-view 3d object detector via historical object prediction.
\newblock In \emph{ICCV}, pp.\  3781--3790, 2023.

\end{thebibliography}
\bibliographystyle{iclr2025_conference}

\clearpage

\appendix

\section{Visualization}

\begin{figure}[h]
    \centering
    \includegraphics[width=\textwidth]{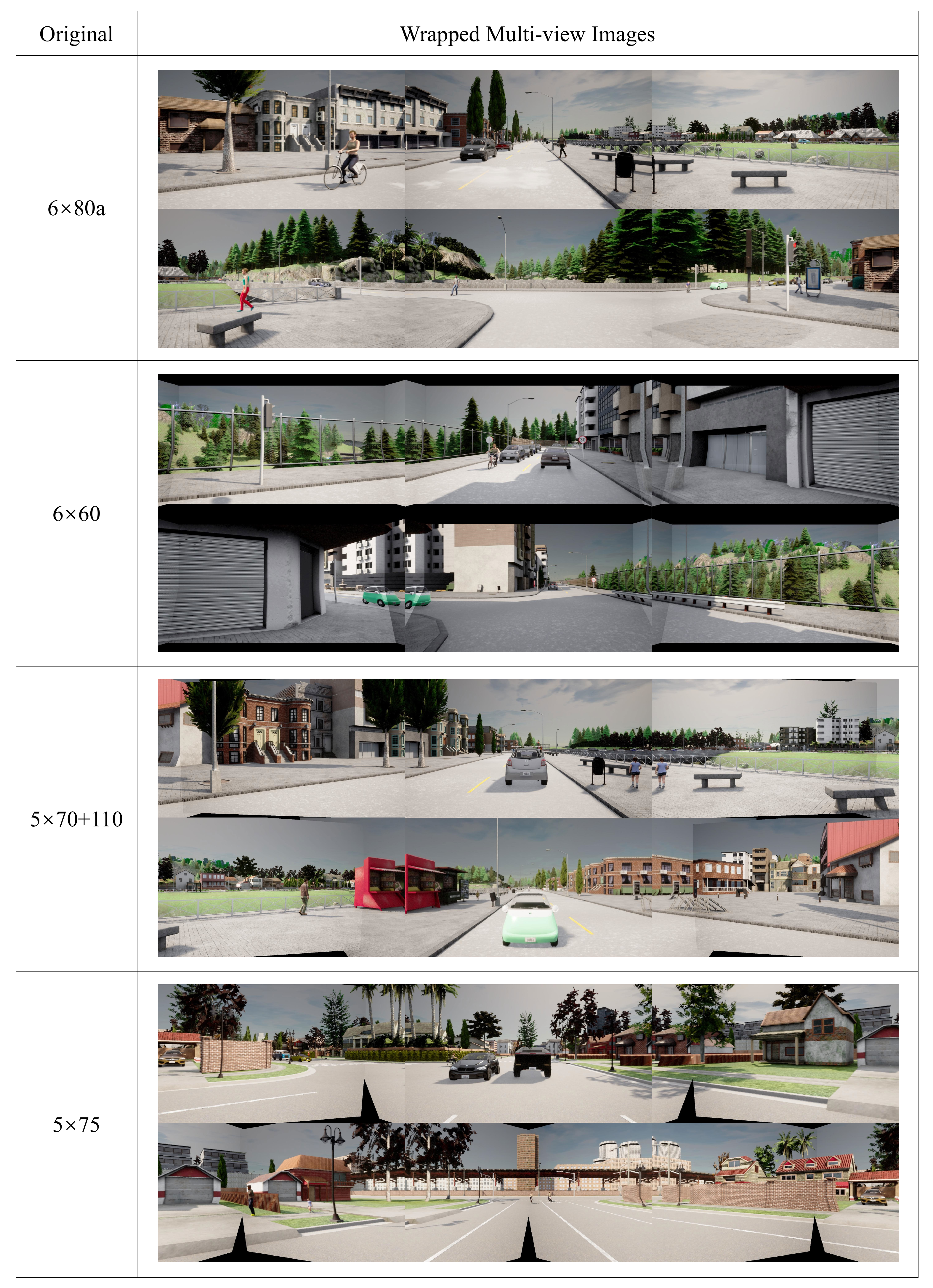}
    \caption{Wrapped Multi-view Images.}
    \label{fig:wrap_figure}
\end{figure}

We present the visualization results of the virtual camera projection in Figure~\ref{fig:wrap_figure}. Overall, the warping from the original view to the virtual view is highly accurate. Only a few areas are not warped because the original cameras lack coverage of those regions.

\end{document}